\documentclass{article}

\PassOptionsToPackage{numbers, compress}{natbib}
\usepackage[preprint]{neurips_2023}

\usepackage[utf8]{inputenc} % allow utf-8 input
\usepackage[T1]{fontenc}    % use 8-bit T1 fonts
\usepackage[pagebackref,breaklinks,colorlinks]{hyperref}
\usepackage{url}            % simple URL typesetting
\usepackage{booktabs}       % professional-quality tables
\usepackage{amsfonts}       % blackboard math symbols
\usepackage{nicefrac}       % compact symbols for 1/2, etc.
\usepackage{microtype}      % microtypography
\usepackage{xcolor}         % colors

\usepackage{amsmath}
\usepackage{amsfonts}
\usepackage{amssymb}
\usepackage{wrapfig}
\usepackage{subcaption}
\usepackage{multirow}
 \usepackage{mathtools} 

\usepackage{verbatim}

\usepackage{anyfontsize}

\usepackage{microtype}
\usepackage{graphicx}
\usepackage{booktabs} % for professional tables
\usepackage{multirow}
\usepackage{amsmath,amssymb}
\usepackage{booktabs}
\usepackage{caption,subcaption}

\usepackage{xcolor}
\definecolor{mygreen}{HTML}{3cb44b}
\definecolor{skyblue}{HTML}{beffff}
\definecolor{lightgreen}{HTML}{90ee90}

\usepackage{color, colortbl}

\definecolor{emerald}{rgb}{0.31, 0.78, 0.37}

\usepackage{tcolorbox}
\usepackage{enumitem}
\setitemize{itemsep=10pt,topsep=0pt,parsep=0pt,partopsep=0pt}
\usepackage{colortbl}

\usepackage{xcolor}
\definecolor{mygreen}{HTML}{3cb44b}
\colorlet{myyellow}{green!10!orange!90!}
\makeatletter

\usepackage{tikz}
\usetikzlibrary{arrows,shapes,snakes,automata,backgrounds,fit,petri}
\usepackage{adjustbox}

\newcommand{\RN}[1]{%
	\textup{\lowercase\expandafter{\it \romannumeral#1}}%
}
\usepackage{tabu}
\newcommand{\ie}[0]{\emph{i.e., }}

\newcommand{\beq}{\vspace{0mm}\begin{equation}}
\newcommand{\eeq}{\vspace{0mm}\end{equation}}
\newcommand{\beqs}{\vspace{0mm}\begin{eqnarray}}
\newcommand{\eeqs}{\vspace{0mm}\end{eqnarray}}
\newcommand{\barr}{\begin{array}}
\newcommand{\earr}{\end{array}}

\usepackage{color, colortbl}
\definecolor{Gray}{gray}{0.93}

\usepackage{lipsum}

\usepackage{pifont}% http://ctan.org/pkg/pifont

\usepackage{makecell}

\usepackage{xcolor,amsmath}
\usepackage[linesnumbered,ruled,vlined]{algorithm2e}
\DontPrintSemicolon

\usepackage{xcolor}
\definecolor{mygreen}{HTML}{3cb44b}

\SetKwComment{Comment}{\color{green!50!black}\# }{}

\SetKwProg{Function}{def}{:}{}

\SetKwProg{For}{for}{:}{}
\SetKwProg{If}{if}{:}{}

\usepackage{xcolor}

\title{Large Multimodal Models: \\Notes on CVPR 2023 Tutorial}

\author{%
  Chunyuan Li \\ 
  Microsoft Research, Redmond \\
  \texttt{\url{https://chunyuan.li}} 
}

\begin{document}

\maketitle

\begin{abstract}
  This tutorial note summarizes the presentation on {\em Large Multimodal Models: Towards Building and Surpassing Multimodal GPT-4}, a part of CVPR 2023 tutorial on {\em Recent Advances in Vision Foundation Models}. The tutorial consists of three parts. We first introduce the background on recent GPT-like large models for vision-and-language modeling to motivate the research in instruction-tuned large multimodal models (LMMs). As a pre-requisite, we describe the basics of instruction-tuning in large language models, which is further extended to the multimodal space. Lastly, we illustrate how to build the minimum prototype of multimodal GPT-4 like models with the open-source resource, and review the recently emerged topics.
\end{abstract}

\begin{figure}[h!]
\centering  
\vspace{-0mm}
\includegraphics[width=0.90\textwidth]{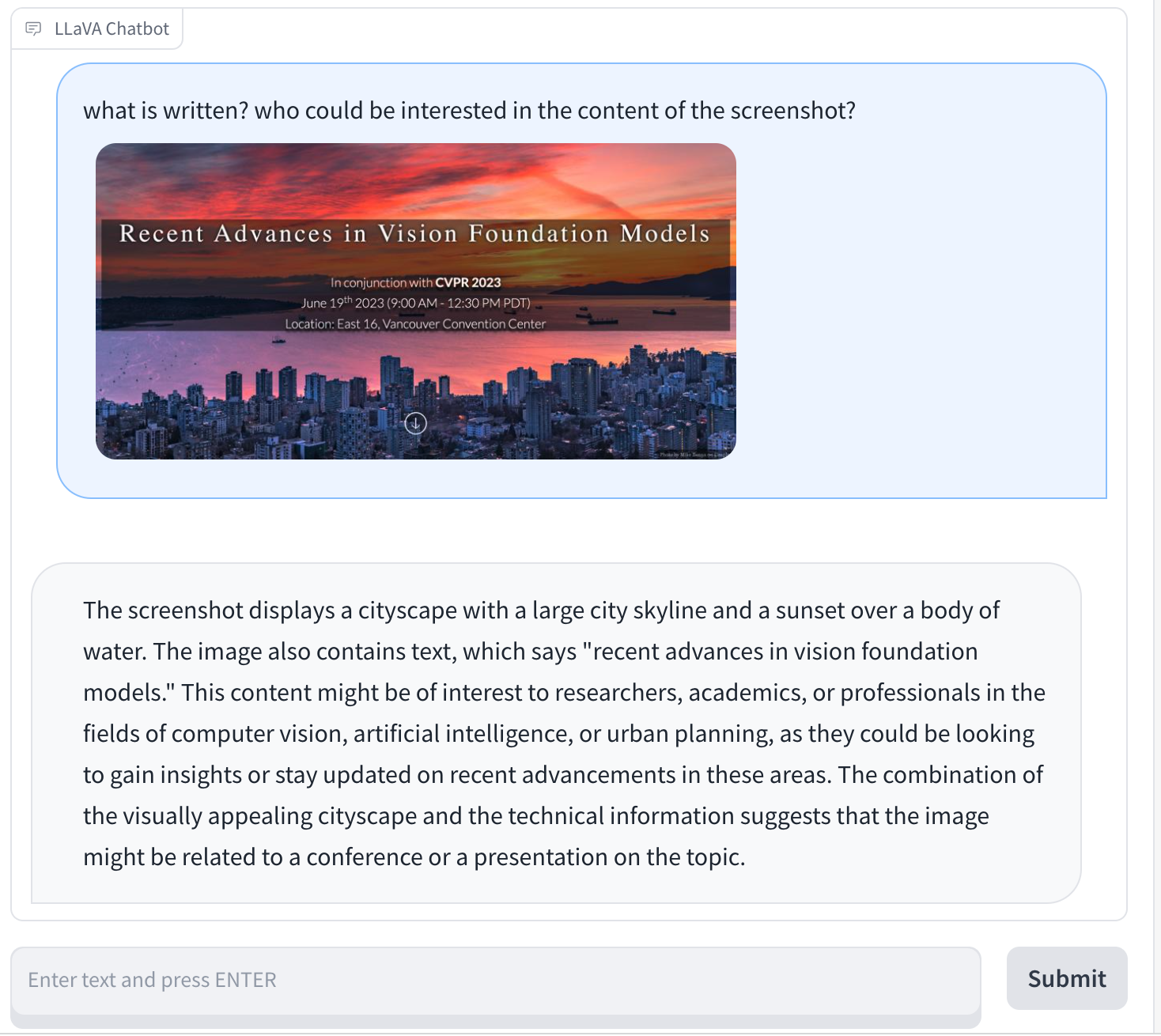}
\vspace{-1mm}
\caption{Visual chat with LMM. Generated by LLaVA: \url{https://llava-vl.github.io/}}
\label{fig:tutorial_screenshot}  
  \vspace{-3mm}
\end{figure}

\newpage
\tableofcontents

\newpage

\section{Prologue}
In view of the rapid assimilation and widespread adoption of OpenAI ChatGPT~\cite{chatgpt}/GPT-4~\cite{gpt4} in contemporary society, there has been a growing interest among academics and researchers to develop open-source large language models (LLMs), and simultaneously explore the extensions into large multimodal models (LMMs)\footnote{Within this manuscript, we will utilize the terms {\em LMM} and {\em multimodal LLM} interchangeably.}. In order to elucidate this popular topic for a broader audience, in the CVPR 2023 tutorial on {\em Recent Advances in Vision Foundation Models}, we have provided a lecture on {\em Large Multimodal Models: Towards Building and Surpassing Multimodal GPT-4}, based on the public materials in the literature. This note summarizes the tutorial presentation and makes it more complete. It gives guided tours through the literature and explain topics to those who seek to learn the areas on LMMs from basics to the advances. It is prepared for audience including graduate students, researchers and professionals that LMMs are outside their specialties, to help them develop perspectives, and identify trends in LMMs in an accessible way. 

\begin{figure}[h!]
\centering  
\vspace{-4mm}
\includegraphics[width=0.99\textwidth]{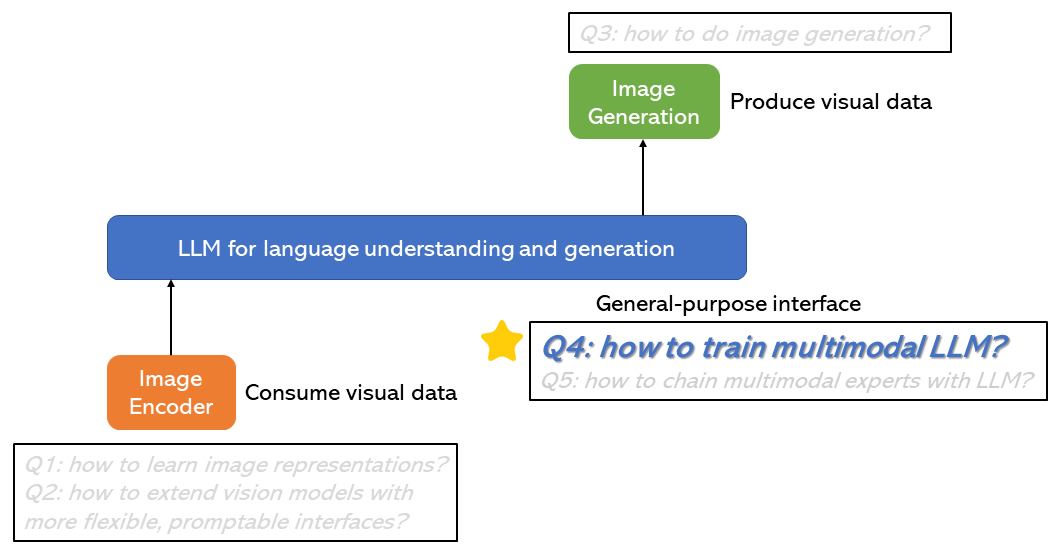}
\vspace{-3mm}
\caption{The focus of this note on large multimodal models, in the context of overall CVPR 2023 Tutorial on {\em Recent Advances in Vision Foundation Models}.}
\label{fig:overall_cvpr2023_tutoral}  
  \vspace{-3mm}
\end{figure}

In the full tutorial, as shown in Figure~\ref{fig:overall_cvpr2023_tutoral}, we have covered the most recent approaches and principles at the frontier of learning and applying vision foundation models, including Q1: Visual and Vision-Language Pre-training; Q2: Generic Vision Interface; Q3: Alignments in Text-to-image Generation; Q4: Large Multimodal Models; and Q5: Multimodal Agents. 

This note focuses on Q4: how to leverage LLM for multimodality, and train LMMs in an end-to-end fashion, so that the models can see and chat. The presentation consists of three parts. To start, we first share background on recent GPT-like large models for vision-and-language modeling in Section~\ref{sec:background}. In the 2nd part, as a pre-requisite, we will introduce the concept of instruction tuning in language domains  in Section~\ref{sec:instruct_tuning_llm}, which empowered ChatGPT. Finally, Section~\ref{sec:instruct_tuning_lmm} covers the last part of the presentation, where we focus on how to build a minimum version of multimodal GPT-4, using LLaVA as a running example. Since LMM is a popular research topic, many new papers have appeared in this  line of research in the past three months, of which we provide a summary, so that the audience may quickly get a picture on what the LMM community has been working on.

The related links of the tutorial presentation on large multimodal models are available at:
\begin{itemize}[leftmargin=7.5mm]
\setlength{\itemsep}{2pt}

\item
{\it Slides}:
\url{https://tinyurl.com/5c2c2mtm}
% \url{https://datarelease.blob.core.windows.net/tutorial/vision_foundation_models_2023/slides/Chunyuan_cvpr2023_tutorial_lmm.pdf}

\item 
{\it YouTube Video}: \url{https://youtu.be/mkI7EPD1vp8}

\item
{\it Bilibili Video}: \url{https://www.bilibili.com/video/BV1Ng4y1T7v3/}

\end{itemize}

For the full information and other parts of the CVPR tutorial, please see
the official website at:
\begin{center}
\vspace{-3mm}
  \url{https://vlp-tutorial.github.io/}
\end{center}

\section{Background}
\label{sec:background}

\subsection{Image-to-Text Generative Models}

\begin{figure}[h!]
\centering  
\vspace{-4mm}
\hspace{-2mm}
\begin{tabular}{p{1.0\textwidth}}
\includegraphics[width=1.00\textwidth]{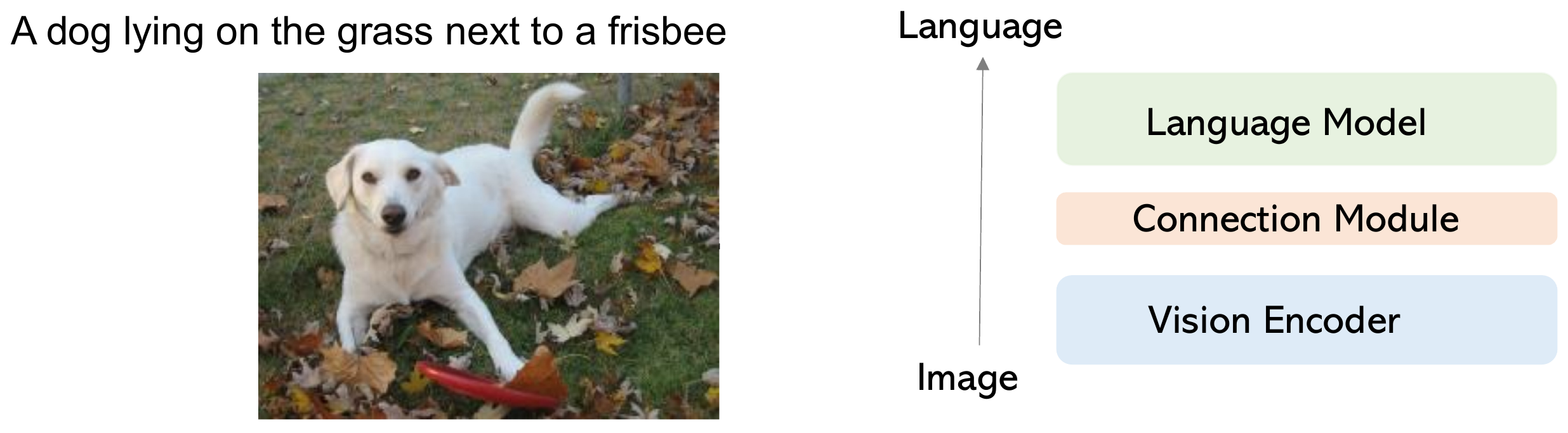} \\
(a) Left: An example of image-to-text generation task; Right: model architecture. \\
\includegraphics[width=1.00\textwidth]{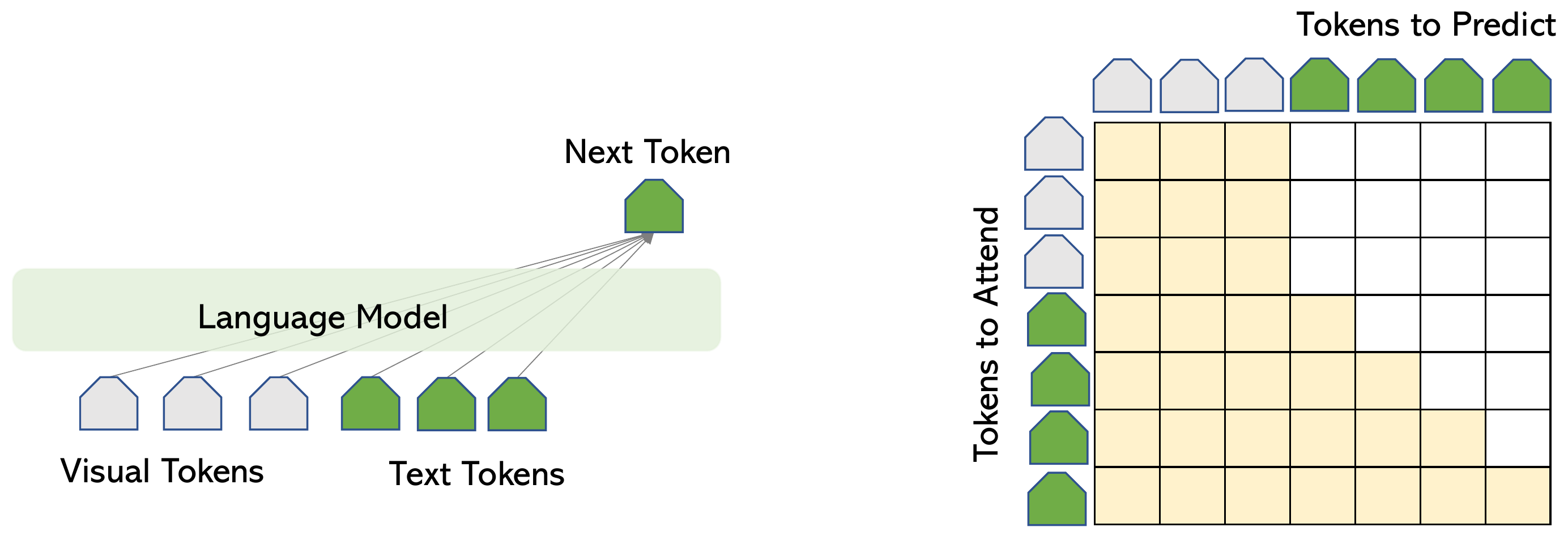} \\
(b) Training objective and attention mask. For each row, the yellow elements indicate that the prediction token attends the tokens on the left. \\
\end{tabular}
\vspace{-0mm}
\caption{Illustration of image-to-text generation task, architecture, and training objective. }
\label{fig:image2text}  
  \vspace{-1mm}
\end{figure}

LMMs in their current form is primarily an image-to-text generative model, which takes images as input, and outputs a text sequence. One example is illustrated in Figure~\ref{fig:image2text} (a) Left. All of the model variants share very similar model architecture and training objective.

\begin{itemize}[leftmargin=7.5mm]
\setlength{\itemsep}{2pt}
\item 
{\bf \it Model Architecture}. As illustrated in Figure~\ref{fig:image2text} (a) Right, the model typically consists of an image encoder to extract visual features, and a language model to decode the text sequence. The vision and language modalities can be optionally connected by trainable connection module. The image encoder and language model can be either trained from scratch or initialized from pre-trained models.

\item
{\bf \it Training Objective}. As illustrated in Figure~\ref{fig:image2text} (b), it typically employs an auto-regressive loss on the output text tokens.
For the attention map in the Transformers~\cite{vaswani2017attention},  image tokens can attend to each other, and the text token depends on and all image tokens and the previous text tokens.
\end{itemize}

\subsection{Case Studies}
We use some known LMMs as examples to illustrate how the network architecture framework can be instantiated in different models, while maintaining the same auto-regressive training objective.

\begin{figure}[t!]
\centering  
\vspace{-4mm}
\hspace{-2mm}
\begin{tabular}{l}
\includegraphics[width=1.00\textwidth]{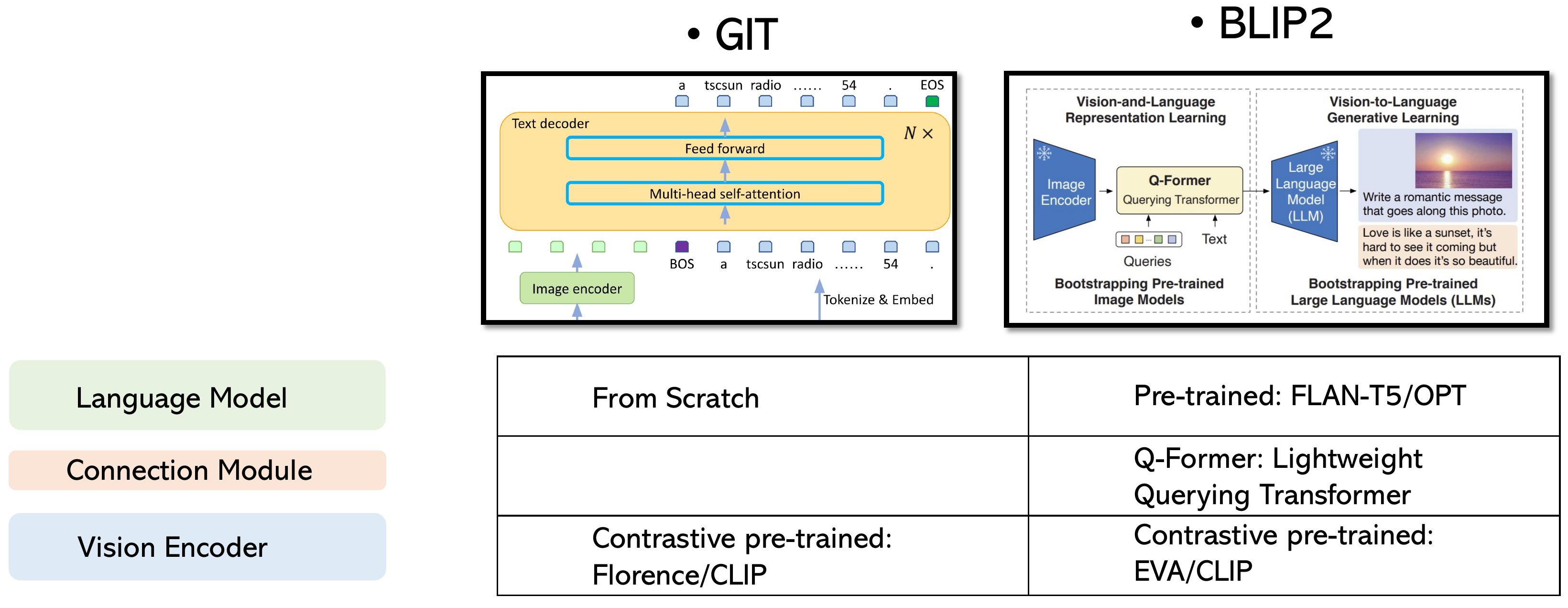} \\
(a) Example 1: LMM with Image-Text Pairs. \vspace{2mm}\\
\includegraphics[width=1.00\textwidth]{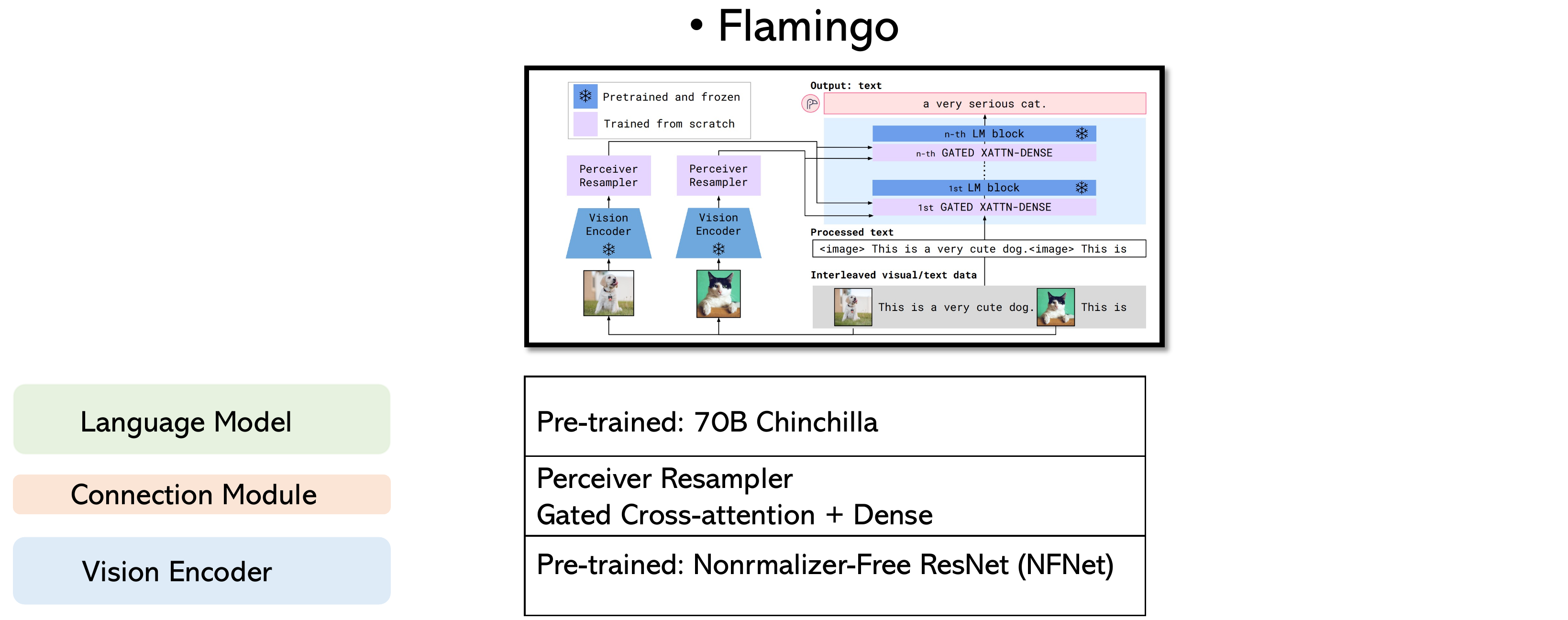} \\
(b) Example 2: LMM with Interleaved Image-Text Data. \\
\end{tabular}
\vspace{-0mm}
\caption{Examples of image-to-text generation models. Image credits are from~\cite{wang2022git,li2023blip,alayrac2022flamingo}.}
\label{fig:image2text_examples}  
  \vspace{-1mm}
\end{figure}

\paragraph{Case Study I: LMM trained with image-text pairwise instances.}
Most LMMs are trained on a large number of image-text pairs, where each training sample is a pair. GIT and BLIP2 are two large models that achieve state-of-the-art (SoTA) performance on many datasets. The comparisons are shown in Figure~\ref{fig:image2text_examples}(a). GIT~\cite{wang2022git} initializes image encoder with constrastive pre-trained Microsoft Florence model, and train a language model from scratch. On the other hand, BLIP2 freezes the weights of pre-trained image and language model, and a train lightweight Q-former. BLIP2~\cite{li2023blip} shows higher sample-efficiency with the bootstrapping training method.

\paragraph{Case Study II: LMM trained with interleaved image-text sequence instances.} We use Flamingo~\cite{alayrac2022flamingo} as example, shown in Figure~\ref{fig:image2text_examples}(b).
It connect the frozen pre-trained image and language models – by adding novel architectural components in between. Specifically, Perceiver Sampler module helps reduce compute complexity, and Gated Transformer module helps stabilize training in the initial stage.
Flamingo is trained on a mixture of complementary large-scale multimodal data coming only from the web, without using any data annotated for machine learning purposes. After this training is done, Flamingo can be directly adapted to vision tasks via simple few-shot learning without any additional task-specific tuning.

\begin{figure}[t!]
\centering  
\vspace{-0mm}
\includegraphics[width=1.00\textwidth]{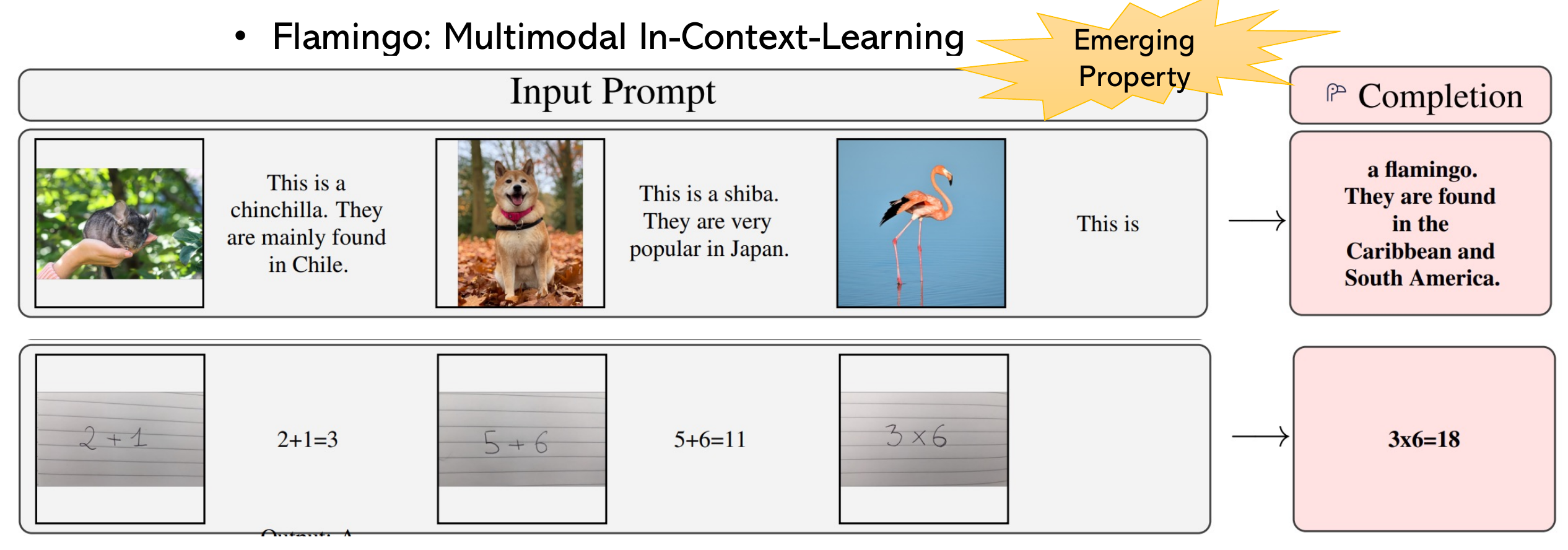} \\
\vspace{-0mm}
\caption{The emerging property of pre-training on web-scale interleaved image-text data: multimodal in-context-learning. Visual examples are from~\cite{alayrac2022flamingo}.}
\label{fig:image2text_mic_emgerging_property}  
  \vspace{-1mm}
\end{figure}

\paragraph{Multimodal In-Context-Learning.}
Beside the SoTA performance on dozens of academic benchmarks, proabably the most appealing aspect of Flamingo is that it exhibits an emerged property: Multimodal In-Context-Learning. Specifically, given a couple of image-text pairs as examples, Flamingo can zero-shot task transfer to new unseen problems, such as solving visual math problems.
This means Flamingo can tackle a number of difficult problems with just a handful of task-specific examples, without any additional training required. 
For example in Figure~\ref{fig:image2text_mic_emgerging_property}, two new tasks are presented to Flamingo. The top row provides two image-text pairs as the context in the prompt, where the text describes the name of the animal in the image, followed by the geographical information of the animal. Flamingo is able to understand the patterns in the task instruction illustrated by the examples, and output the corresponding information for a new image. In the bottom row, the text first shows the optical character recognition (OCR) result of the image, followed by the arithmetic result. Flamingo learns the task instruction illustrated in the multimodal context, outputs the correct answer for a new math problem in the image.
Therefore, Flamingo is generally considered as the GPT-3 moment~\cite{brown2020language} in the multimodal domain.

\subsection{OpenAI Multimulti GPT4 and Research Gaps}
\begin{figure}[h!]
\centering  
\vspace{-4mm}
\includegraphics[width=1.00\textwidth]{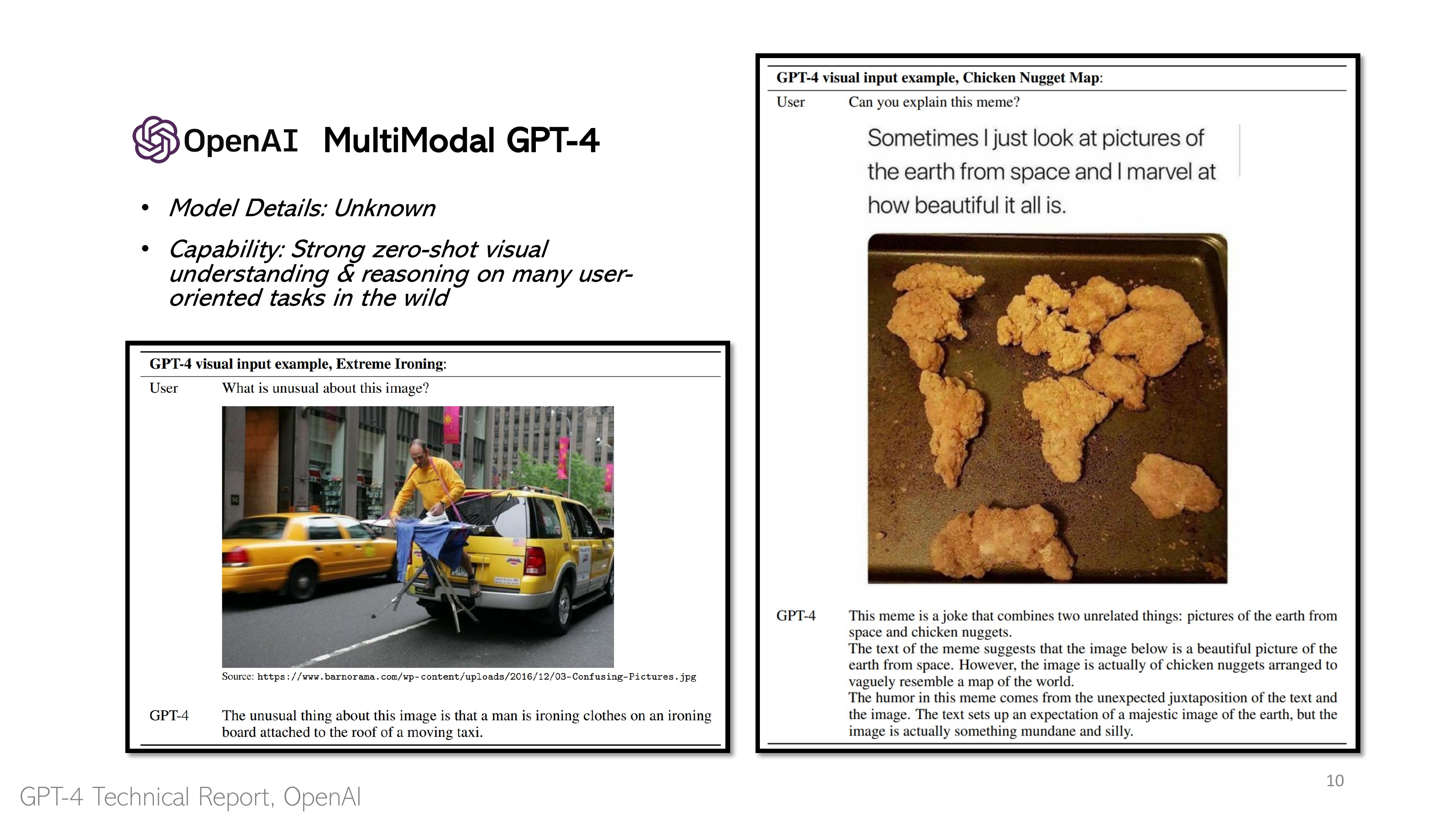} \\
\vspace{-0mm}
\caption{OpenAI MultiModal GPT-4. Visual examples are from~\cite{gpt4}.}
\label{fig:gpt4_examples}  
  \vspace{-1mm}
\end{figure}

In March 2023, OpenAI released GPT-4~\cite{gpt4}, with impressive capability in visual understanding and reasoning. Though the model details are unknown, there is no doubt that GPT4 enables many new scenarios, based on the examples highlighted the technique report. For instance, two popular visual examples are illustrated in Figure~\ref{fig:gpt4_examples}. The first one identifies  the uncommon visual region and exhibits strong complex reasoning performance. The second one recognizes text in the image and captures the mere across image-text. For a while, the research community had no clue how this new ability is achieved (probably because they are not tightened to any established academic tasks/datasets), but all are determined that these are exciting results.
It naturally raise a question: How can we build Multimodal GPT-4 like models?

\begin{figure}[t!]
\centering  
\vspace{-4mm}
\includegraphics[width=1.00\textwidth]{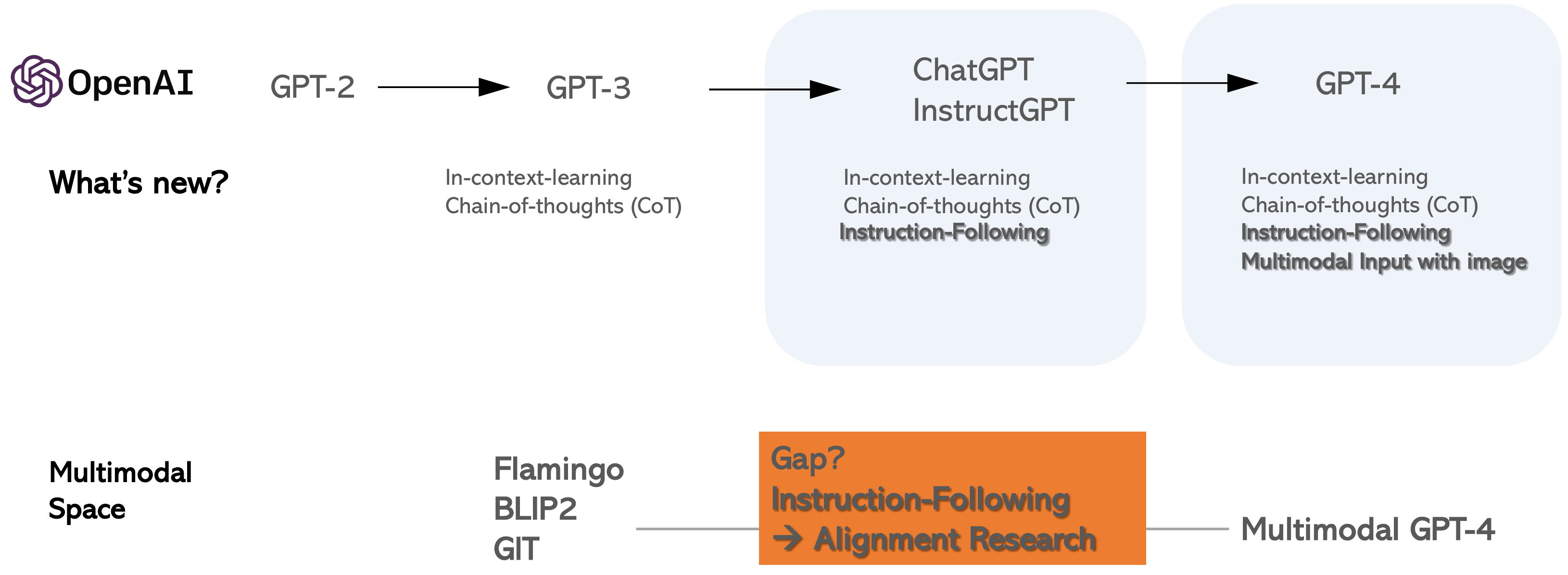} \\
\vspace{-0mm}
\caption{Recap on Language Modeling: OpenAI LLM development history. The unique properties for each generation model is highlighted, from which the research gap is revealed for LMM.}
\label{fig:lmm_research_gap}  
  \vspace{-1mm}
\end{figure}

To answer it, we start to review the big models from OpenAI, by highlighting the most appealing properties for each model in Figure~\ref{fig:lmm_research_gap}. There are several key observations:
$(i)$
GPT-2~\cite{radford2019language} is the auto-regressive counterpart in the BERT era~\cite{devlin2018bert} for the paradigm of pre-training then fine-tuning. Compared with GPT-2, GPT-3~\citep{brown2020language} is a 175B model trained on web-scale text corpus, which exhibits two emerging properties with a frozen model: in-context-learning~\citep{brown2020language} and chain-of-thoughts (CoT) reasoning~\citep{wei2022chain}.. 
This means, without any additional training required,  the model can tackle a wide range of new problems with just a few task-specific examples and by properly prompting it step-by-step, respectively. 
It further leads to the paradigm from fine-tuning model weights to prompting frozen models, where the latter shows higher generality and lower adaptation cost in task transfer.
$(ii)$ ChatGPT and InstructGPT~\cite{ouyang2022training} shows the importance of instruction-following and alignment with human intents for LLMs, by fine-tuning the base language model GPT-3/GPT-3.5 on high quality instruction-following data, and improving them with a reward model via reinforcement learning with human feedback.
$(iii)$ GPT-4 not only improves the language ability of previous models, but also allows visual signals as additional input for understanding and reasoning. We see that the newer generation model  maintains/improves the existing properties of the previous ones, and enable new properties. 

In another words, from GPT-3 to GPT-4, we see two new properties: instruction-following and multimodal input.
This reveals the gap between existing LMMs such as Flamingo and multimodal GPT-4: how to perform instruction-following and alignment research in the multimodal space. and thus the focus of this tutorial \&  note.

% =====================================
% \pagebreak
\newpage

\section{Pre-requisite: Instruction Tuning in Large Language Models}
\label{sec:instruct_tuning_llm}
Note that instruction-following is a notion originated in natural language processing (NLP). To study the intuition and gain a full picture of the history, we revisit instruction tuning with LLMs.

\subsection{Instruction Tuning}
\label{sec:what_is_instruct_tuning_llm}

\begin{figure}[h!]
\centering  
\vspace{-4mm}
\hspace{-2mm}
\begin{tabular}{p{1.0\textwidth}}
\includegraphics[width=1.00\textwidth]{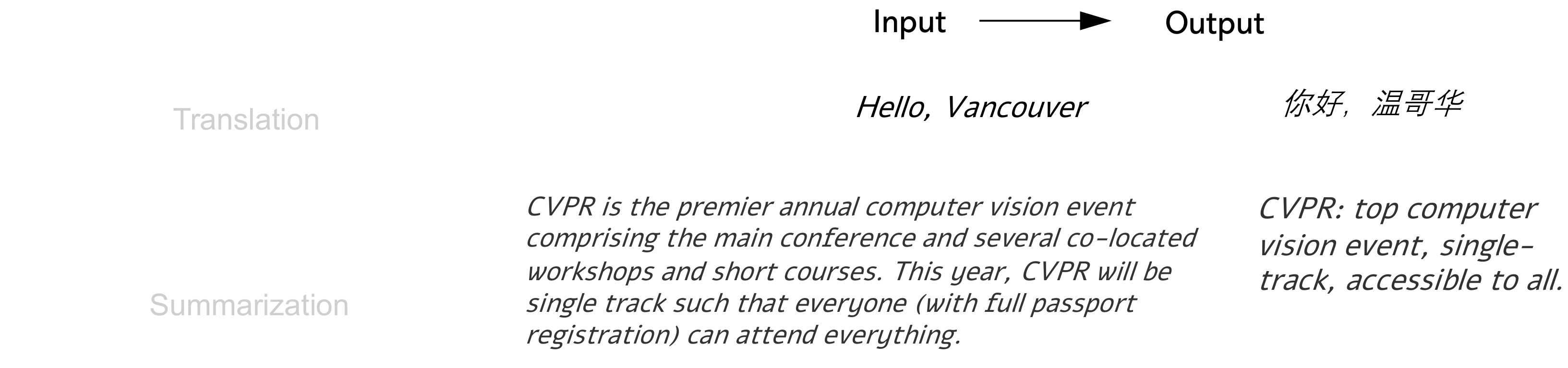} \\
(a) Training: Implicit task instructions in traditional language data. \vspace{2mm}\\
\includegraphics[width=1.00\textwidth]{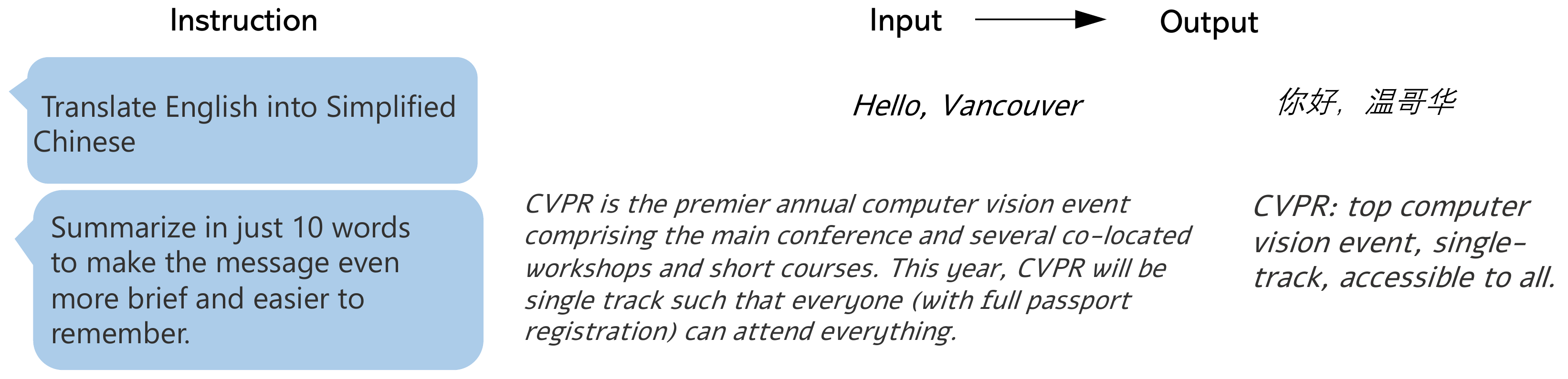} \\
(b) Training: Explicit task instructions in instruct language data. \\
\includegraphics[width=1.00\textwidth]{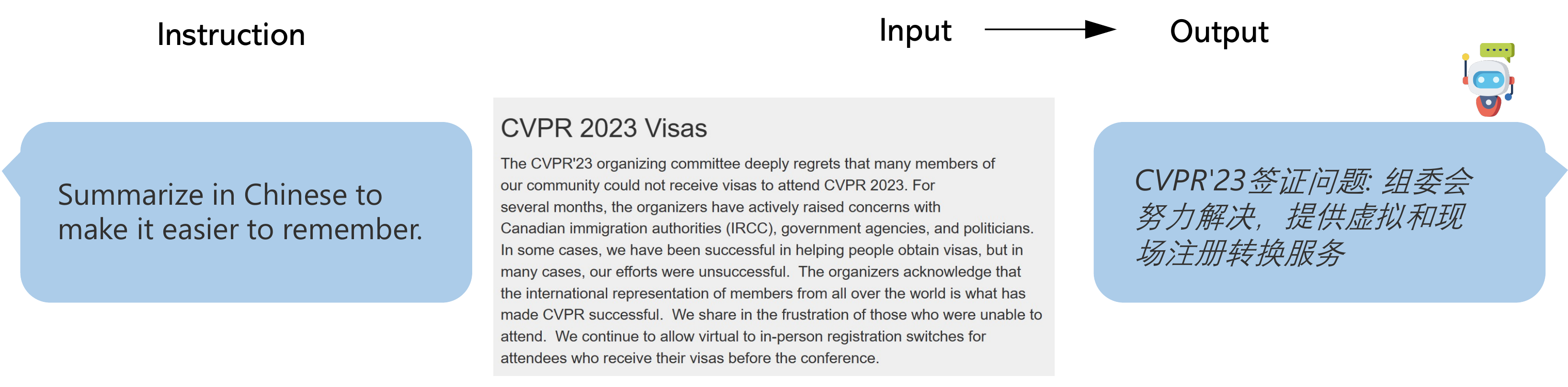} \\
(c) Inference: Explicit task instructions in instruct language data. The English meaning of the output is {\em "CVPR'23 visa issue: organizing committee works to solve and  provide virtual and in-person registration switch services."
} \\
\end{tabular}
\vspace{-0mm}
\caption{Examples of task instructions in traditional and instruct language data, respectively.}
\label{fig:task_instruction_lm}  
  \vspace{-1mm}
\end{figure}

\paragraph{Traditional Language Data.}
As a typical data instance in NLP, seq2seq representation is quite common for many language tasks: each data instance consists of two parts: sequence as the input and sequence as the output. We provide two examples in Figure~\ref{fig:task_instruction_lm} (a). Without any task instruction specified, we know they are translation and summarization tasks, respectively.

This seq2seq representation is also how NLP community used to use their data. Task instructions are implicit. Based on each data domain, individual models are trained, or sometimes multi-tasking over multiple data domain without specifying the task instructions. When such models are trained, they are  hard to generalize to new tasks in a zero-shot fashion, because the models do not learn the skill to understand the task instruction, and have no ability to distinguish and generalize what task to perform in the testing stage.

\paragraph{Instruct Language Data.}
Instead, recently researchers start to explicitly add task instructions in the model training, as shown in Figure~\ref{fig:task_instruction_lm} (b). Interestingly, the task instructions of most NLP tasks can be expressed in natural language as well. It leads a new data format: instruction-input-output triplets. Based on the new format, one single model can be trained, multi-tasking with specified instructions. Since models have observed many task instructions and many instances for each task in training, it is natural and easy for the models to generalize to new tasks by task composition in the inference stage.

For example, in the evaluation stage, a new task that require both summarization and translation is provided in Figure~\ref{fig:task_instruction_lm} (c). Though the model has never seen this new task in training, it observes individual task basis, and learn to perform on new tasks. Note that we humans are always creating new tasks in our daily life, and presumably these new tasks would never been observed by models. It is thus appealing if a model is able to solve thousands of new tasks in the wild in without training. This is partially why ChatGPT is becoming popular and prevalent quickly.

\subsection{Self-Instruct and Open-Source LLMs}
\label{sec:what_is_instruct_tuning_llm}

How can we collect a diverse set of high-quality instruction-following data? There are two general schemes. One is human-human interaction, where humans (task providers) provide the annotation statement and requirements, based on which another group of humans complete the annotation tasks. such a scheme is typically cost and time consuming. The other scheme is human-machine interaction, where similarly humans provide the annotation statement and requirements, but it is now the machines/models that complete the annotation tasks. 

\begin{figure}[h!]
\centering  
\vspace{-4mm}
\includegraphics[width=1.00\textwidth]{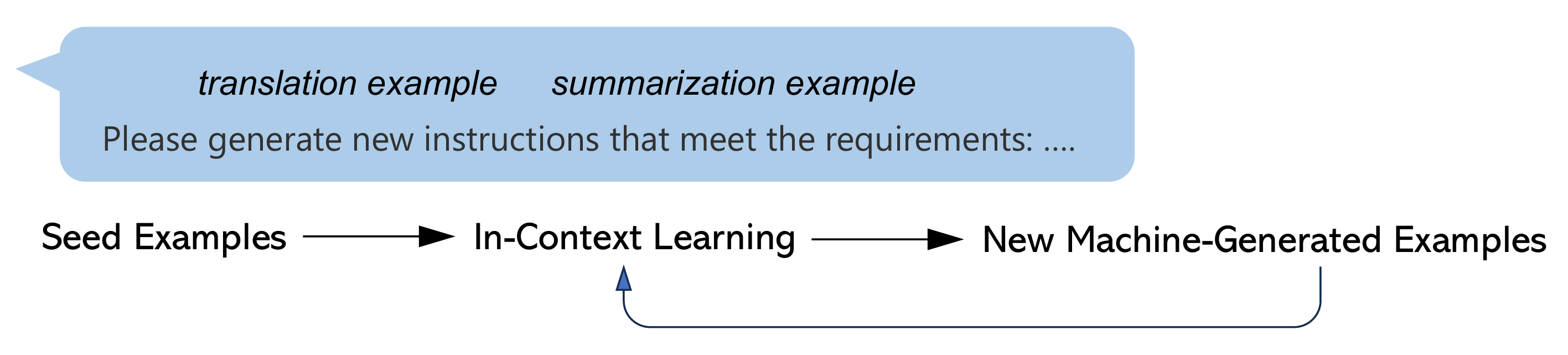} \\
\vspace{-0mm}
\caption{Self-instruct pipeline.}
\label{fig:self_instruct}  
  \vspace{-1mm}
\end{figure}

To enable LLMs to follow natural language instructions 
and complete real-world tasks, researchers have been exploring methods of instruction-tuning of LLMs. This is implemented by either fine-tuning the model on a wide range of tasks using human-annotated prompts and feedback~\citep{ouyang2022training}, or supervised finetuning using public benchmarks and datasets augmented with manually or automatically generated instructions~\citep{wang2022benchmarking}. Among these methods, Self-Instruct tuning~\citep{wang2022self} is a simple and effective method of aligning LLMs to human intent, by learning from instruction-following data generated by SoTA teacher LLMs. It turns out that the line of instruction-tuning research has produced effective means to improve the zero and few-shot generalization abilities of LLMs. Self-instruct leverages the in-context-learning ability of LLM. The pipeline is illustrated in Figure~\ref{fig:self_instruct}. Humans create a few examples (\ie seed examples) as the context, and ask LLM such as GPT-3 or GPT-4 to create more instruct and responses that follows the requirements stated in the prompt. The machine-generated instruction-following data can be further selected to construct with the prompt for in-context-learning in the next data generation iteration. The procedure iterates till a given number of samples are collected.
Due to the relatively lower cost and higher response speed of API calls (compared with human annotations), self-instruct is becoming more favorable in the research community.

\begin{figure}[h!]
\centering  
\vspace{-0mm}
\includegraphics[width=1.00\textwidth]{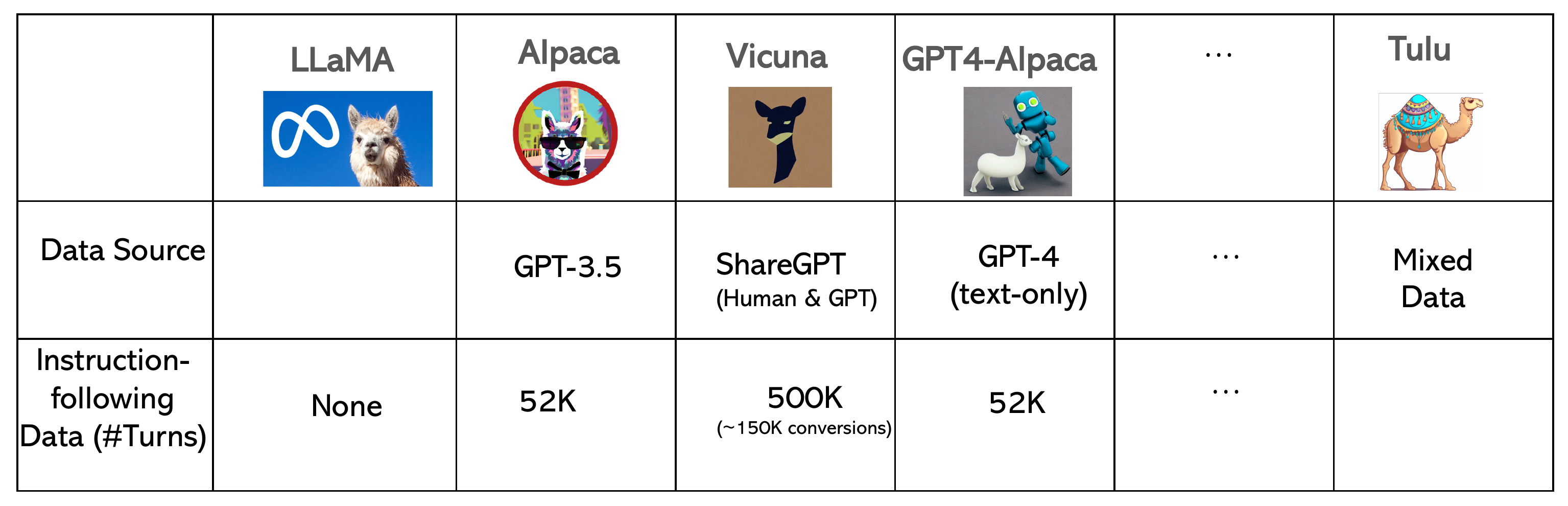} \\
\vspace{-0mm}
\caption{Model examples of the LLaMA family.}
\label{fig:llama_family}  
  \vspace{-1mm}
\end{figure}

\paragraph{Open-Source LLMs: LLaMA Family.}
The open-source community has witnessed a surge of open LLM.
The success of ChatGPT~\citep{chatgpt} and GPT-4~\citep{gpt4} offers tremendous opportunities to improve open-source LLMs using instruction-tuning. 
Figure~\ref{fig:llama_family} compares several open-source instruction tuned LLMs. 
LLaMA~\citep{touvron2023llama} is a series of open-sourced LLMs, which match the performance of proprietary LLMs such as GPT-3. To teach LLaMA to follow instructions, Self-Instruct tuning has been quickly adopted given its superior performance and low cost. For example, to name a few early attempts in this line of research, Stanford Alpaca~\citep{alpaca} uses 52K instruction-following samples generated by GPT-3.5, while Vicuna~\citep{vicuna} uses around 500K high-quality instruction-following samples (150K conversions) between user and GPT~\citep{sharegpt}. To advance the SoTA of instruction-tuning for LLMs, GPT-4 is utilized as the teacher to generate the responses for the Alpaca instructions~\cite{peng2023instruction}. Many papers have been proposed to improve the instruction-following data to improve the model alignment quality in chat. For a comprehensive review, we suggest the readers to refer the recent paper~\cite{wang2023far}, where a LLM Tulu is trained on a mix of several high-quality instruct data, and comprehensive comparisons are conducted across multiple benchmarks.

\begin{figure}[t!]
\centering  
\vspace{-0mm}
\includegraphics[width=1.00\textwidth]{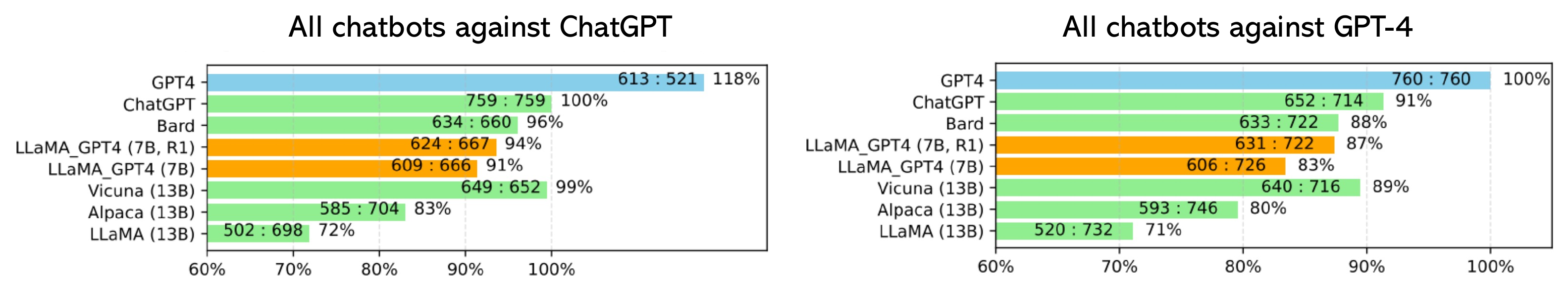} \\
\vspace{-0mm}
\caption{Model performance evaluated on Vicuna-80 questions.}
\label{fig:llm_performance}  
  \vspace{-1mm}
\end{figure}

\paragraph{Quick Assessment of LLM Chatbots.} 
To study the quality of LLM Chatbots, We consider {\it Vicuna-Instructions-80}\footnote{\footnotesize \url{https://github.com/lm-sys/FastChat/blob/main/fastchat/eval/table/question.jsonl}}~\citep{vicuna}, a dataset with 80 challenging questions that baseline models find challenging. Beside generic instructions, there are 8 categories, including knowledge, math, Fermi, counterfactual, roleplay, generic, coding, writing, common-sense.
To quantitatively compare the performance, we ask GPT-4 to rate the response from score 1 to 10 for any two given chatbots, then compute the relative score. The results are shown in Figure~\ref{fig:llm_performance}. 
Surprisingly, it turns out this evaluation metric is quite consistent across different settings. The open-source LLaMA family seem performing closely to SoTA proprietary Chatbots. 
% 
% So, how we are doing in the multimodal space.

\paragraph{Further Discussions.} There are several important topics on LLMs that we have not covered in the tutorial presentation, but are worthwhile future exploring.

\begin{itemize}[leftmargin=7.5mm]
\setlength{\itemsep}{2pt}
\item 
{\bf \it Data-centric AI}. We emphasize that the developmet of these open-source LLM projects is data-centric~\cite{mazumder2022dataperf}, rather than model-centric, so that we hope readers could align the perspective when discussing the topic. As the training objective and network architectures are becoming similar and even identical on GPT-like projects, the key differential factor is data. For example, behaviors of the aforementioned LLMs are determined by the instruction tuning data. 

\item
{\bf \it False Promise?} There is a debate that the open LLMs could catch up with the proprietary LLMs is a false promise~\cite{gudibande2023false}. 
To align the discussions, we argue that there are two distinctive abilities for LLMs: the instruction-following ability to know which task to perform, and massive knowledge storage to complete the task with quality. Imitation models are good at the former, by mimicking ChatGPT's style but not its factuality. They authors in~\cite{gudibande2023false} conclude that there exists a substantial capabilities gap between open and closed LMs that, with current methods, can only be bridged using an unwieldy amount of imitation data or by using more capable base LMs. They also advocate that the highest leverage action for improving open-source models is to tackle the difficult challenge of developing better base LMs. However, unfortunately the resources to train such base LMs are only available in a few industry labs, and the formulas to train the base LMs is largely well explored. It seems more promising for most academic research labs to explore the opportunities in alignment research with affordable resources, or explore the techniques to reduce the compute the barriers.

\item
{\bf \it Base LLMs}. Developing more capable or commercial usable LLMs is of great value. Besides LLaMA, the open-source community has developed several capable base LLMs such as OpenLLaMA~\citep{openlm2023openllama}, MPT~\citep{MosaicML2023Introducing} and Falcon~\citep{refinedweb}, or released the training recipe~\citep{together2023redpajama}.
\end{itemize}

\pagebreak
\newpage

\section{Instructed Tuned Large Multimodal Models}
\label{sec:instruct_tuning_lmm}

In this tutorial, we illustrate how to build the minimum prototype of multimodal GPT4 with open-source resources. Specially, we use LLaVA~\citep{liu2023visual} as the running example, a similar idea is also proposed in its co-current work miniGPT-4~\citep{zhu2023minigpt4}.

\subsection{Open-Source Prototypes: LLaVA / MiniGPT4}
\label{sec:llava}

The research in the multimodal space has often been inspired by the latest advances in NLP in recent years. One successful recipe is to keep asking what would happen if the most intriguing and successful NLP ideas are borrowed for the vision-and-language community. 
We are leveraging the self-instruct idea from the language domain. The unique challenge with self-instruct is that there is no strong multimodal teacher available yet. How can we use language model such as language-only GPT-4 to create multimodal instruction following data.

\begin{figure}[h!]
\centering  
\vspace{-4mm}
\hspace{-2mm}
\begin{tabular}{l}
\includegraphics[width=1.00\textwidth]{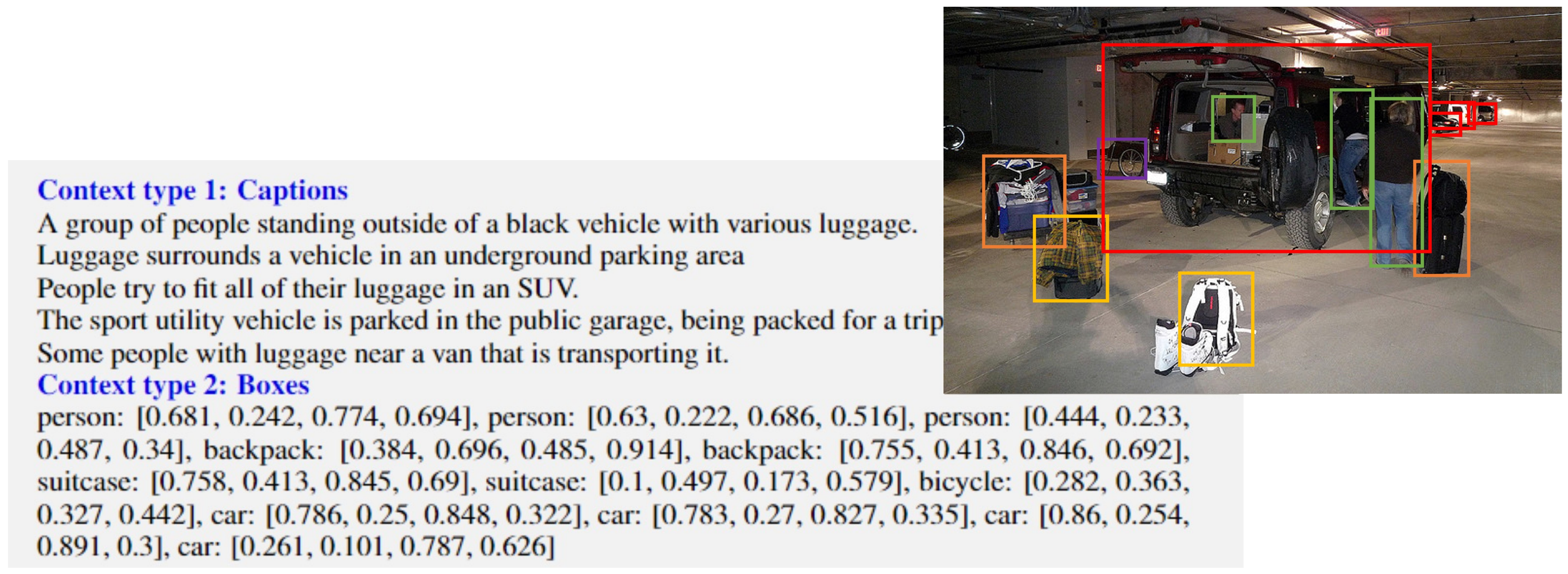} \\
(a) The sequence representation of the image data. \vspace{2mm}\\
\includegraphics[width=0.8\textwidth]{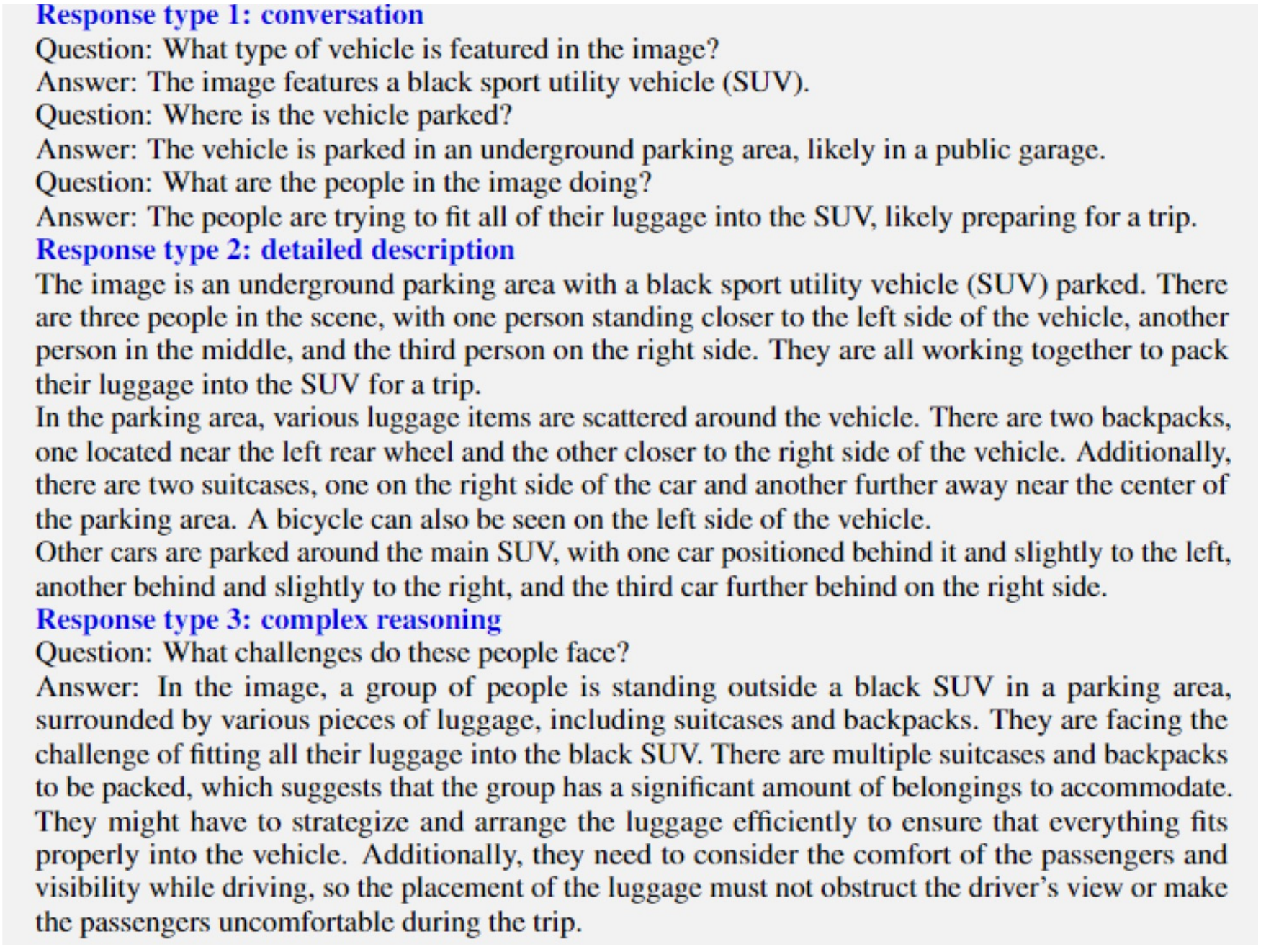} \\
(b) The three types of instruction-following data for the given image. \\
\end{tabular}
\vspace{-0mm}
\caption{Examples of multimodal instructional-following data. Image credits from~\cite{liu2023visual}.}
\label{fig:multimodal_instruction_data}  
  \vspace{-1mm}
\end{figure}

\subsubsection{Data Creation}
Instead of directly feed images into OpenAI GPT, we use their symbolic sequence representations shown in Figure~\ref{fig:multimodal_instruction_data} (a). In LLaVA, the caption and boxes are considered, due to the following reasons: (1) it is empirically found that GPT-4 can understand them well, in contrast that ChatGPT has a difficult time in understanding the box data. (2) they are important to represent the image as informative as possible.

As exemplified in Figure~\ref{fig:multimodal_instruction_data} (b), three types of instruction-following data are considered: multi-turn conversations so that users can chat with bot, detailed description so that long response can be generated from the bot; Lastly, complex reasoning, this is more about the implication of the image, rather than the image content. For example, “what challenge do these people face” in this image? The image is about a SUV in the parking area, while the challenge is how the luggage can be packed into the SUV due to the tight space in the car. In total, 158K samples are collected.

To summarize, the trick is that whatever tasks one wants to the model to perform in the serving stage, it is important to create the corresponding instruction-following for the training.

\begin{figure}[t!]
\centering  
\vspace{-0mm}
\includegraphics[width=0.34\textwidth]{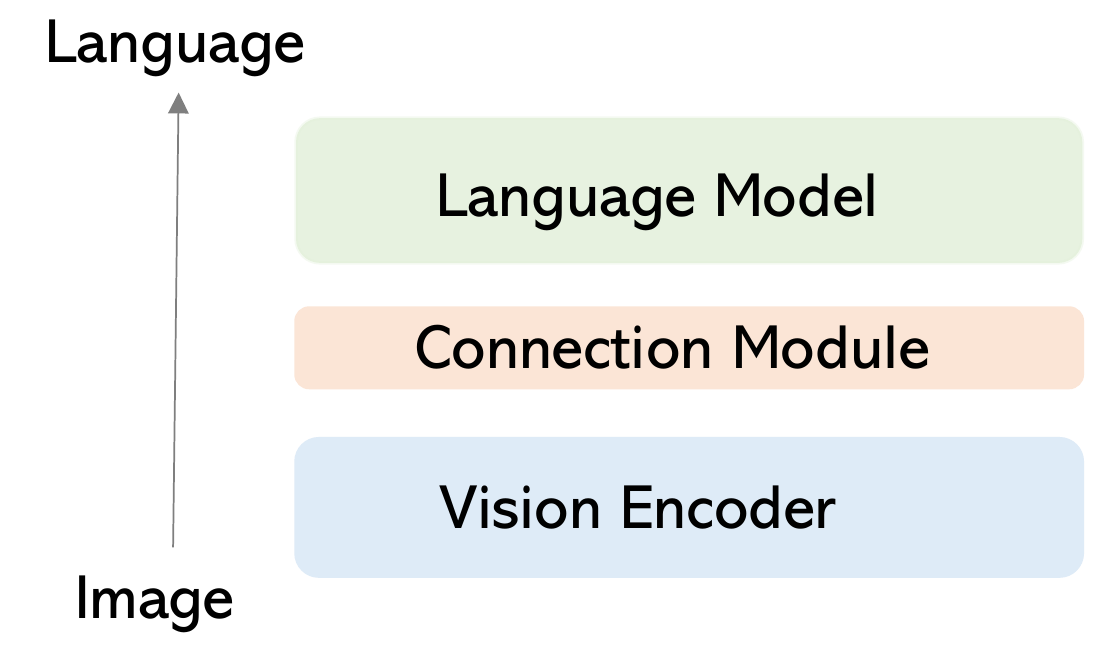} ~
\includegraphics[width=0.6\textwidth]{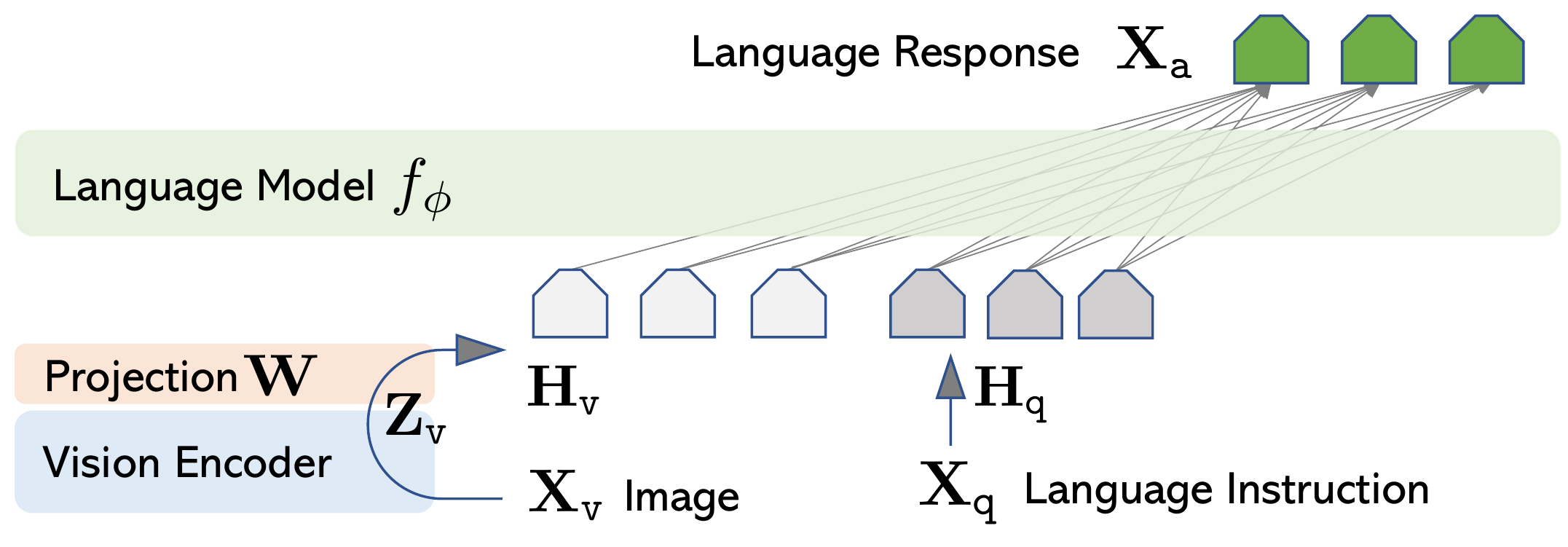} \\
\vspace{-0mm}
\caption{Network architecture: Left: General LMM; Right: LLaVA. Image credits from~\cite{liu2023visual}.}
\label{fig:llava_arch}  
  \vspace{-1mm}
\end{figure}

\subsubsection{Network Architecture and Training}
As illustrated in Figure~\ref{fig:llava_arch}, the LLaVA network architecture is an instantiation of the general image-to-text generative model framework introduced in Section~\ref{sec:background} and Figure~\ref{fig:image2text}. Specifically, LLaVa connects pre-trained CLIP ViT-L/14 visual encoder~\citep{radford2021learning} and large language model Vicuna~\citep{vicuna}, using a simple projection matrix. A two-stage instruction-tuning procedure is considered:

\begin{itemize}[leftmargin=7.5mm]
\setlength{\itemsep}{2pt}
\item 
{\bf \it Stage 1: Pre-training for Feature Alignment.} Only the projection matrix is updated, based on a subset of CC3M~\citep{sharma2018conceptual}. The only task is image captioning.

\item
{\bf \it Stage 2: Fine-tuning End-to-End.} Both the projection matrix and LLM are updated for two different use scenarios.

\end{itemize}

\subsubsection{Performance}
\paragraph{Performance on Visual Chat: Towards building multimodal GPT-4 level chatbot.}. LLaVA is fine-tuned on the generated multimodal instruction-following data, which contains a diverse set of task instruction and response for daily user-oriented applications. It is empirically found that fine-tuning the linear projection layer only is sufficient for the chat demo/scenarios, though it requires longer training time.

\begin{figure}[h!]
\centering  
\vspace{-0mm}
\includegraphics[width=0.9\textwidth]{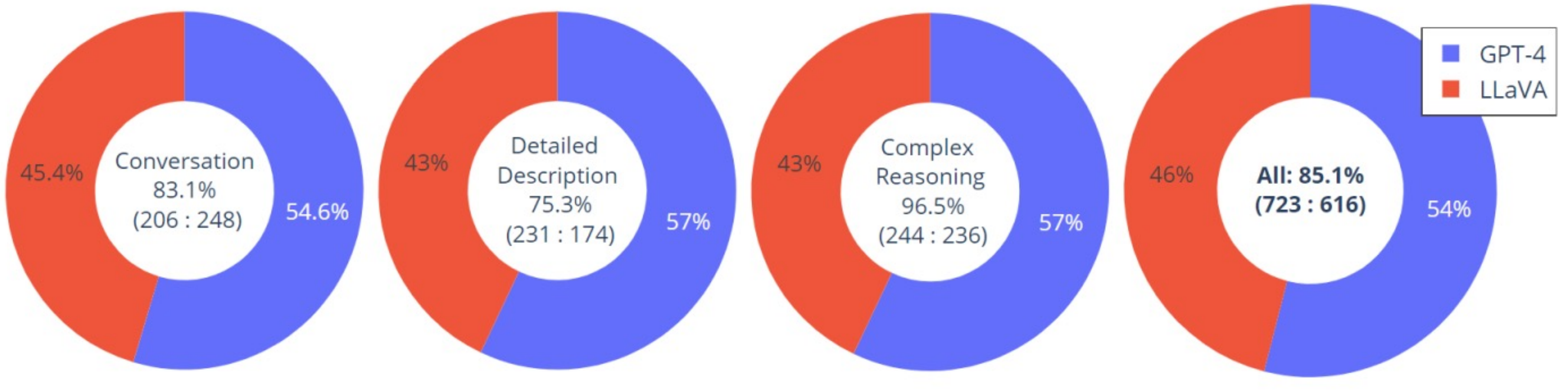} \\
\vspace{-0mm}
\caption{Visual chat performance: LLaVA vs GPT-4. Image credits from~\cite{liu2023visual}.}
\label{fig:visual_chat_performance}  
  \vspace{-1mm}
\end{figure}

An evaluation dataset with 30 unseen images is constructed: each image is associated with three types of instructions: conversation, detailed description and complex reasoning. This leads to 90 new language-image instructions, on which we test LLaVA and GPT-4, and use GPT-4 to rate their responses from score 1 to 10. The summed score and relative score per type is reported in Figure~\ref{fig:visual_chat_performance}. Overall, LLaVA achieves 85.1\% relative score compared with GPT-4, indicating the effectiveness of the proposed self-instruct method in multimodal settings.

More examples are shown in Table~\ref{tab:visual_example_ironing} and Table~\ref{tab:visual_example_chichken}, respectively.

\begin{table}
  \begin{minipage}{0.99\textwidth}
\centering  
\vspace{-4mm}
\scalebox{0.88}{
\begin{tabular}{l p{12.5cm} }
\toprule
 \multicolumn{2}{l}{\bf Visual input example, Extreme Ironing:}  \\
\midrule
&  \includegraphics[height=4.5cm]{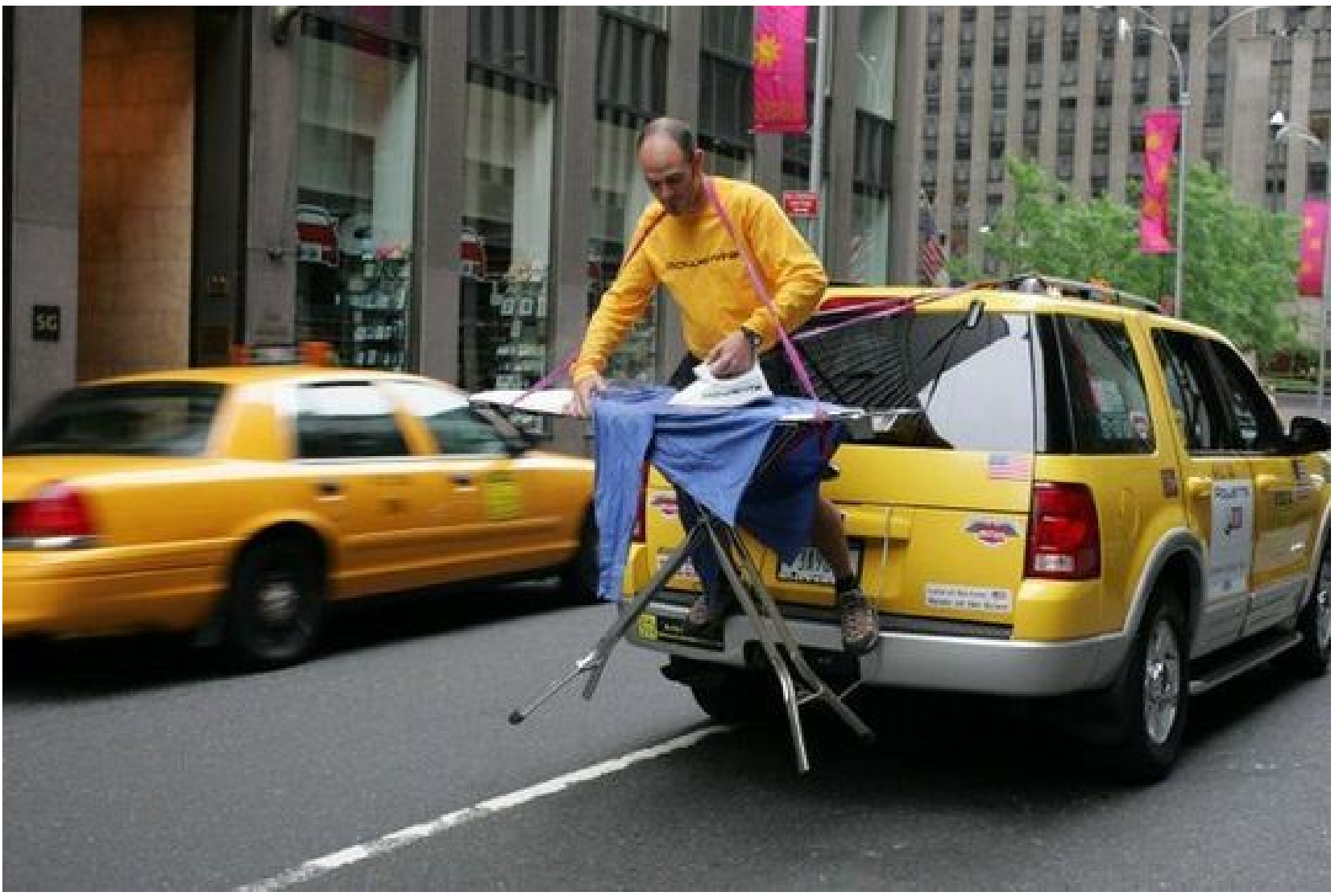} \\
& {\tiny Source:~\url{https://www.barnorama.com/wp-content/uploads/2016/12/03-Confusing-Pictures.jpg}} \\

User & What is unusual about this image? \\
LLaVA & The unusual aspect of this image is a man ironing clothes on the back of a minivan or van. This is not a typical place to perform this activity, as one would usually iron clothes in a more stationary and safe location, such as a home, using a regular ironing board. The scene depicted in the image is peculiar as it involves a makeshift ironing setup on a vehicle, which can be both unsafe and unconventional. Additionally, it is not clear how the man is able to maintain balance and stability while ironing clothes in such an unstable environment.
\\
% \midrule
 & [{\it Start a new conversation, and clear the history}] \\
User & What’s happening in the scene? \\
LLaVA & The image depicts a man standing on top of a yellow SUV in a busy city street. He is holding a portable ladder, seemingly ironing clothes while standing on the vehicle. Around the scene, there are other cars, a traffic light, and a fire hydrant. Additionally, there are two other people in the vicinity, one close to the man on the car and another further away. The man on the ladder appears to be attracting attention \textcolor{blue}{due to his unconventional choice of ironing his clothes on top of a moving car}. The city street around him is bustling with activity, adding to the unique nature of the scene.
\\
\midrule
User & What is unusual about this image? \\
GPT-4~\cite{gpt4} &  The unusual thing about this image is that a man is ironing clothes on an ironing
board attached to the roof of a moving taxi.
\\ \midrule
User & What is unusual about this image? \\
MiniGPT-4 & The image shows a man standing on the back of a yellow taxi cab, holding a piece of clothing in his hand. The taxi is parked on a busy street with other cars passing by. The man is wearing a yellow shirt and has a look of determination on his face. The image is unusual because it shows a person standing on the back of a taxi, which is not a common sight in most cities.
\\ \midrule
User & What is unusual about this image? \\
BLIP-2 & a man is sitting on the back of a yellow cab
\\ \midrule
User & What is unusual about this image? \\
OpenFlamingo & The man is drying his clothes on the hood of his car.
\\
\bottomrule
\end{tabular}
}
\vspace{1mm}
\captionof{table}{Example prompt from GPT-4 paper~\cite{gpt4} to compare visual reasoning and chat capabilities. Compared to BLIP-2~\cite{li2023blip} and OpenFlamingo~\cite{anas_awadalla_2023_7733589}, LLaVA accurately follows the user's instructions, instead of simply describing the scene. LLaVA offers a more comprehensive response than GPT-4. Even when merely asked to describe the image, LLaVA identifies atypical aspects of the image. Table credits from~\cite{liu2023visual}.}
\vspace{-7mm}
\label{tab:visual_example_ironing}  
  \end{minipage}
\end{table}

\begin{table}[h!]
  \begin{minipage}{0.99\textwidth}
\centering
\scalebox{0.88}{
\begin{tabular}{l p{12.5cm} }
\toprule
 \multicolumn{2}{l}{\bf Visual input example, Chicken Nugget Map:}  \\
\midrule
&  \includegraphics[height=4.5cm]{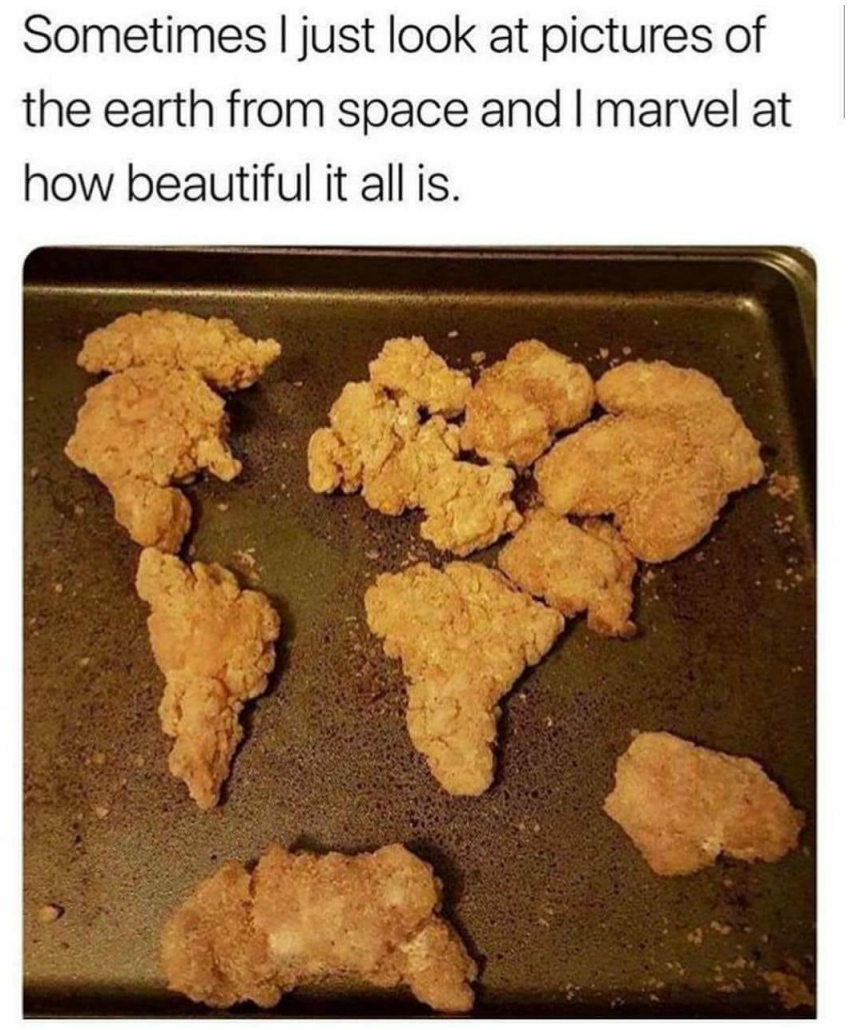} \\

User & Can you explain this meme in detail? \\
LLaVA & The meme in the image is a creative and humorous take on food, with a focus on chicken nuggets, as the center of the universe. The meme begins with the phrase "Sometimes I just look at pictures of the Earth from space and I marvel at how beautiful it all is..." The main image features a tray of cooked chicken nuggets, with the chicken pieces positioned to resemble the continents and islands on a world map.
The meme playfully suggests that the chicken nuggets represent the Earth, and the various locations depicted in the photo are actually chicken nugget versions of different places. The punchline of the meme appears in smaller text at the bottom, reading "I mean, it's not the real Earth, but how beautiful it is all is." This lighthearted and entertaining meme combines food and humor, using the image of chicken nuggets to create a funny and imaginative representation of the world.
\\
\midrule
User & Can you explain this meme? \\
GPT-4~\cite{gpt4} &  This meme is a joke that combines two unrelated things: pictures of the earth from
space and chicken nuggets.
The text of the meme suggests that the image below is a beautiful picture of the
earth from space. However, the image is actually of chicken nuggets arranged to
vaguely resemble a map of the world.
The humor in this meme comes from the unexpected juxtaposition of the text and
the image. The text sets up an expectation of a majestic image of the earth, but the
image is actually something mundane and silly.
\\ \midrule
User & Can you explain this meme in detail? \\
MiniGPT-4 & This meme is a picture of a pan full of fried chicken. The chicken is cooked and appears to be ready to be served. The caption reads, “Sometimes it’s just like looking at pictures from space and marveling at how beautiful it all is.” This meme is likely meant to be humorous and playful, as it compares the beauty of space to the simple pleasure of eating fried chicken. It is likely meant to be a lighthearted and funny way to express appreciation for both the wonders of space and the simple pleasures of food.
\\ \midrule
User & Can you explain this meme in detail? \\
BLIP-2 & sometimes i just look at pictures of the earth from space and marvel how beautiful it is
\\ \midrule
User & Can you explain this meme in detail? \\
OpenFlamingo & It's a picture of a chicken nugget on the International Space Station. 
\\
\bottomrule
\end{tabular}
}
\vspace{2mm}
\captionof{table}{Example prompt comparing LLaVA, GPT-4, BLIP-2, and OpenFlamingo's visual reasoning capabilities in understanding the humor. LLaVA and GPT-4 both explain the meme and its humor, while GPT-4 produces a more concise answer. Table credits from~\cite{liu2023visual}.}
\label{tab:visual_example_chichken}  
  \end{minipage}
  \vspace{-0mm}
\end{table}

\begin{wrapfigure}{r}{0.53\textwidth}
  \begin{minipage}{0.53\textwidth}
\centering  
\vspace{-10mm}
\includegraphics[width=0.90\textwidth]{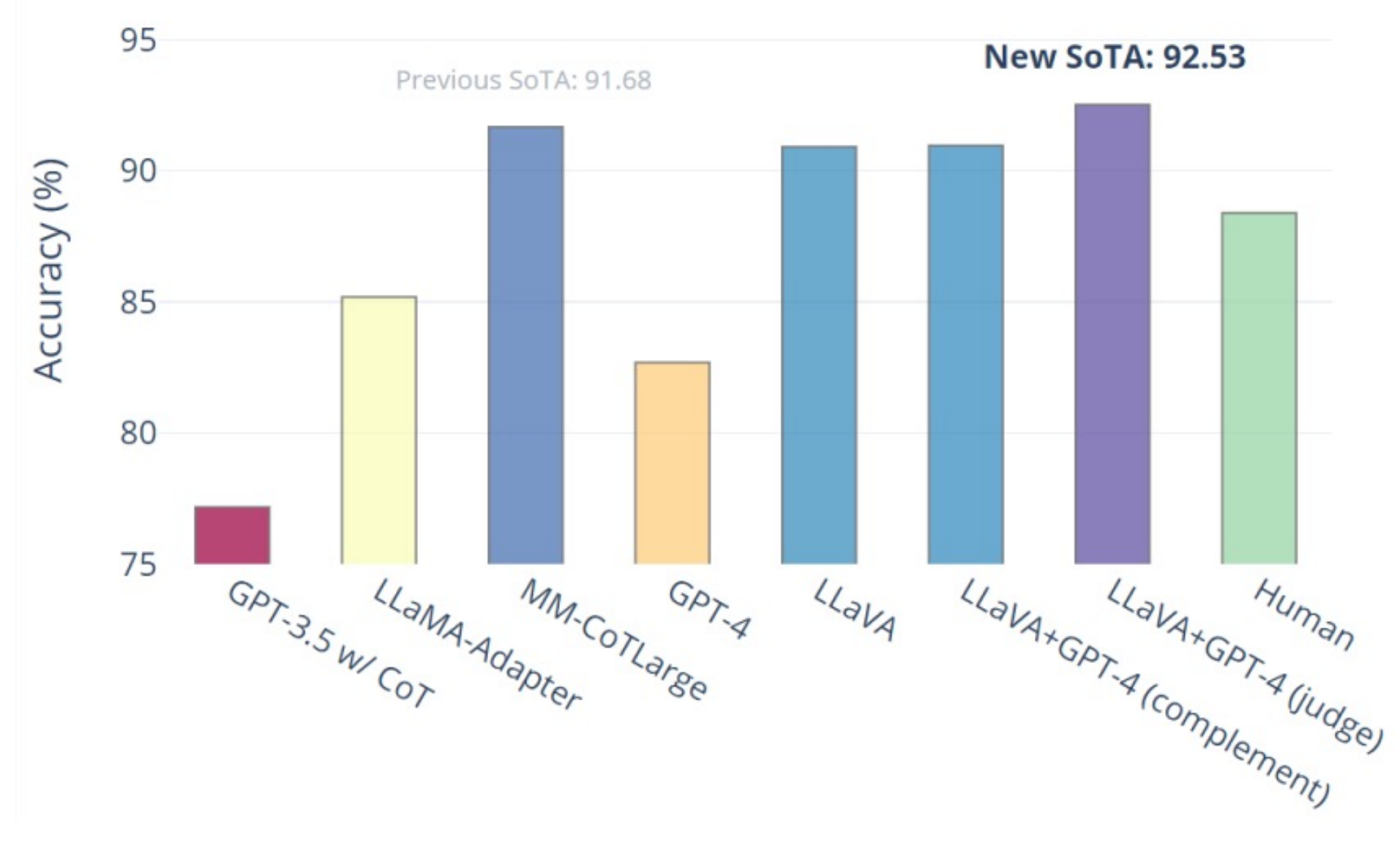}
\vspace{-7mm}
\captionof{figure}{Science QA.}  
\label{fig:science_qa}  
  \end{minipage}
\end{wrapfigure}
\paragraph{Performance on Science QA: New SoTA with the synergy of LLaVA with GPT-4.}
LLaVA is fine-tuned on a multimodal reasoning dataset in the science domain~\citep{lu2022learn}.  
In Figure~\ref{fig:science_qa}, LLaVA alone achieves 90.92\%. We use the language-only GPT-4 as the judge, to predict the final answer based on its own previous answers and the LLaVA answers. This ``GPT-4 as judge'' scheme yields a new SoTA 92.53\%.

\begin{figure}[h!]
\centering  
\vspace{-6mm}
\hspace{-2mm}
\begin{tabular}{c}
\includegraphics[width=0.65\textwidth]{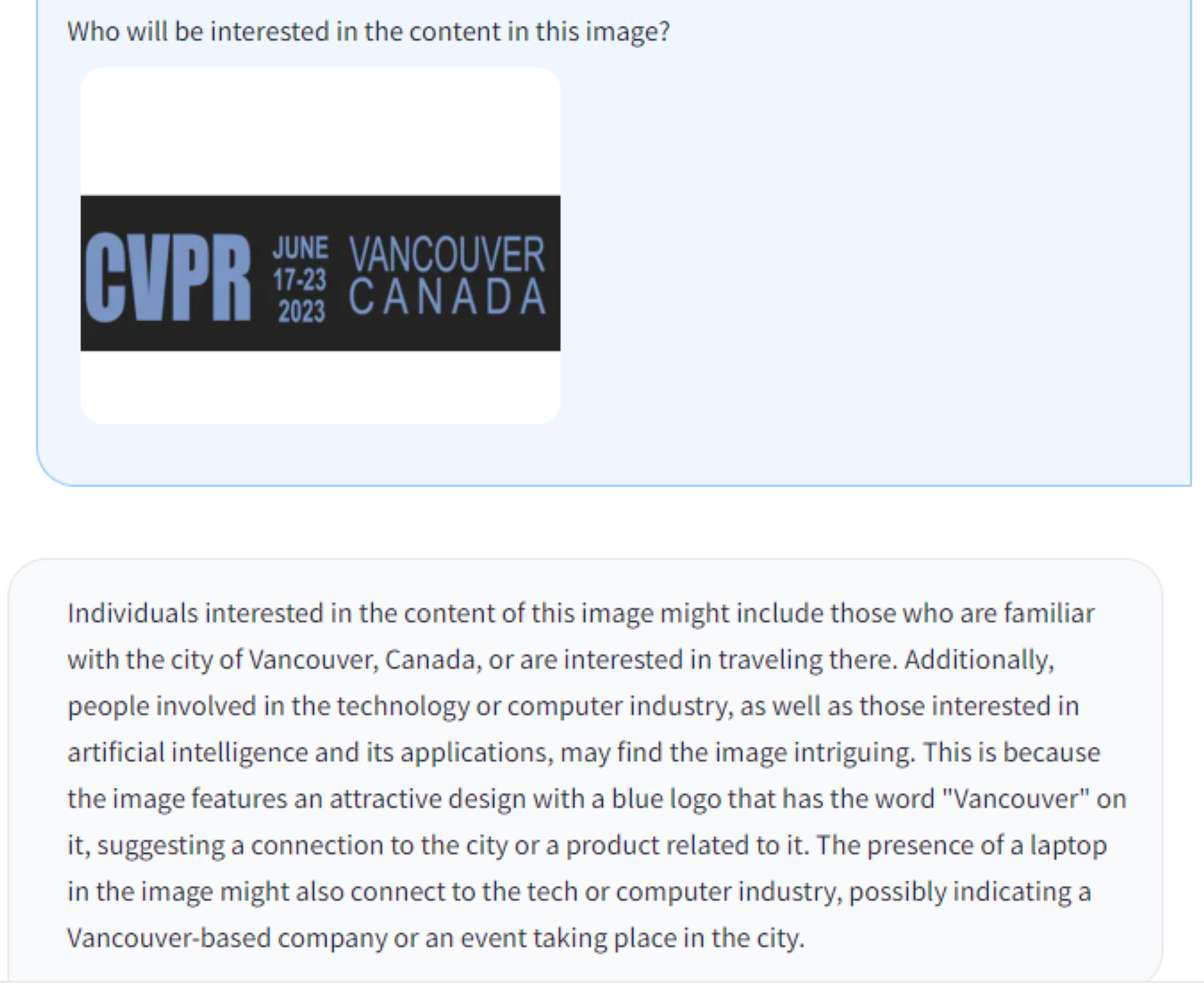} \\
(a) CVPR 2023 logo. \url{https://cvpr2023.thecvf.com/}  \vspace{2mm}\\
\includegraphics[width=0.65\textwidth]{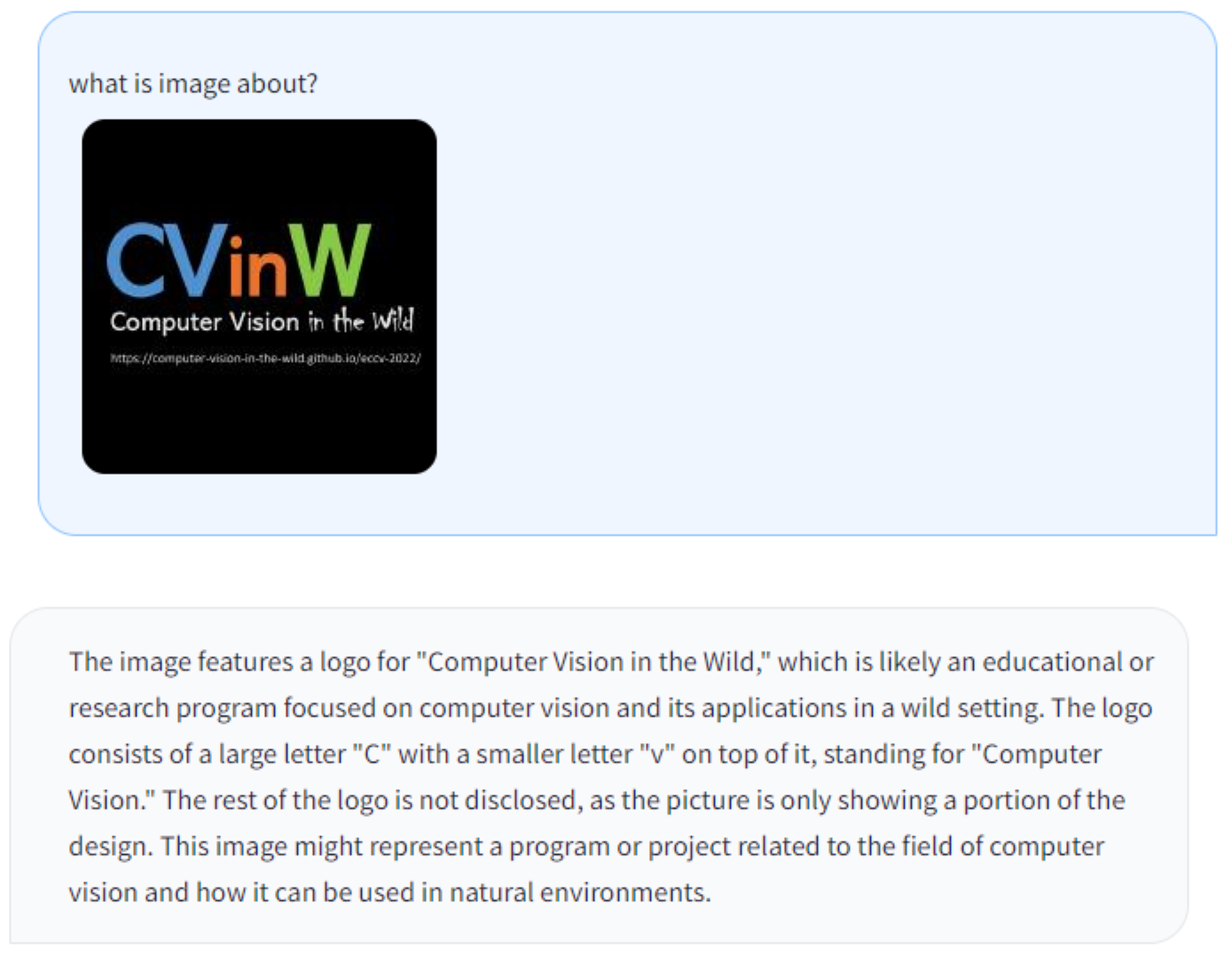} \\
(b) CVinW workshop logo. \url{https://github.com/Computer-Vision-in-the-Wild} \\
\includegraphics[width=0.65\textwidth]{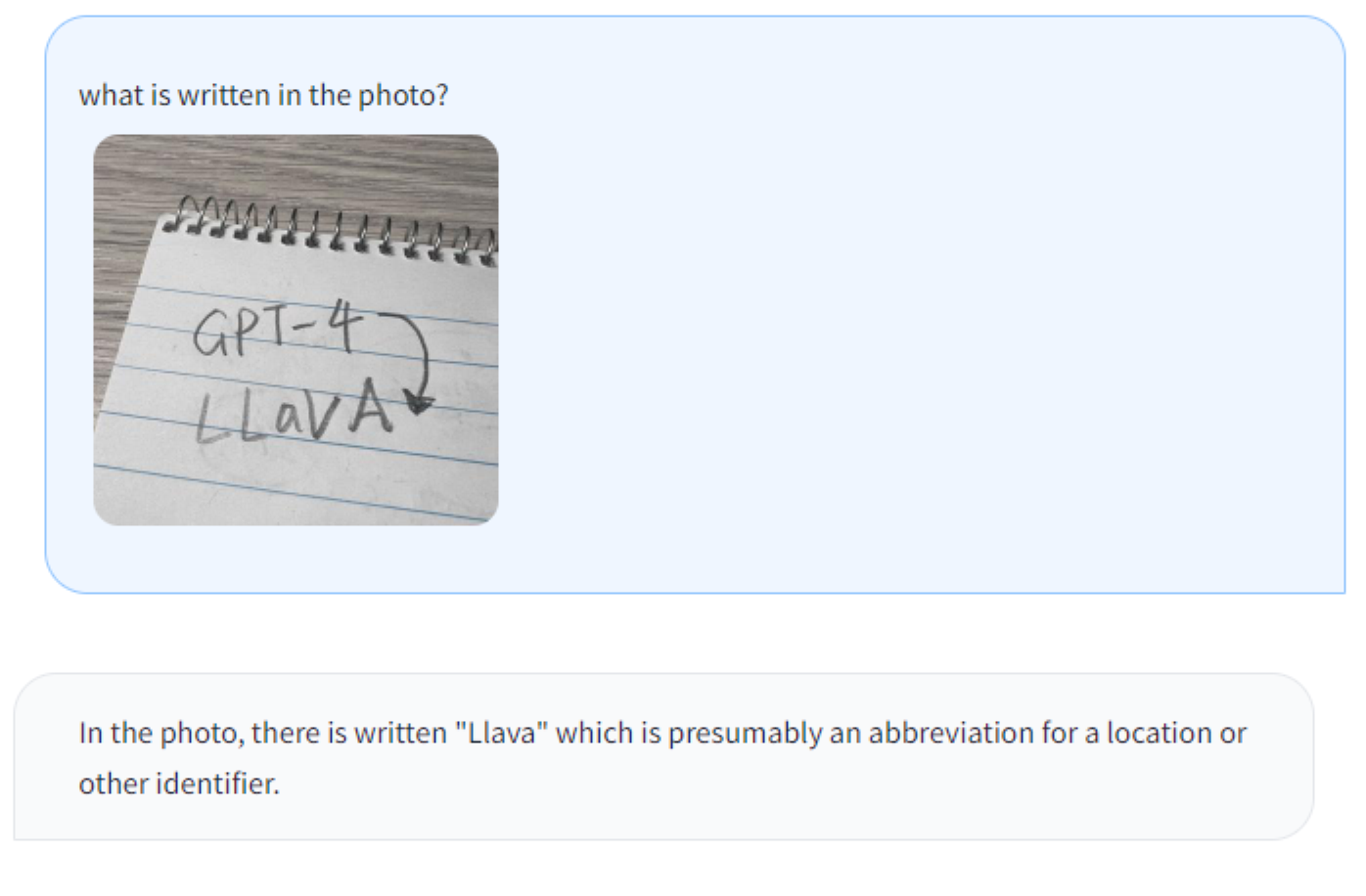} \\
(c) Hand-written ``LLaVA''. \url{https://llava-vl.github.io/} \\
\end{tabular}
\vspace{-0mm}
\caption{Examples that LLaVA is able to recognize and reasons with text in the images in the wild.}
\label{fig:ocr_examples_llava}  
  \vspace{-1mm}
\end{figure}

\paragraph{Performance on OCR in the wild: An emerging property.} LLaVA has never been explicitly trained on OCR data, \ie images that contains text from the corresponding caption. Surprisingly, the model show strong zero-shot OCR task transfer ability in the wild. Some examples are shown in Figure~\ref{fig:ocr_examples_llava}.

\newpage
\subsection{Emerging Topics}
\label{sec:emerging_topics}

\begin{figure}[h!]
\centering  
\vspace{-3mm}
\hspace{-2mm}
\begin{tabular}{p{1.0\textwidth}}
\includegraphics[width=1.0\textwidth]{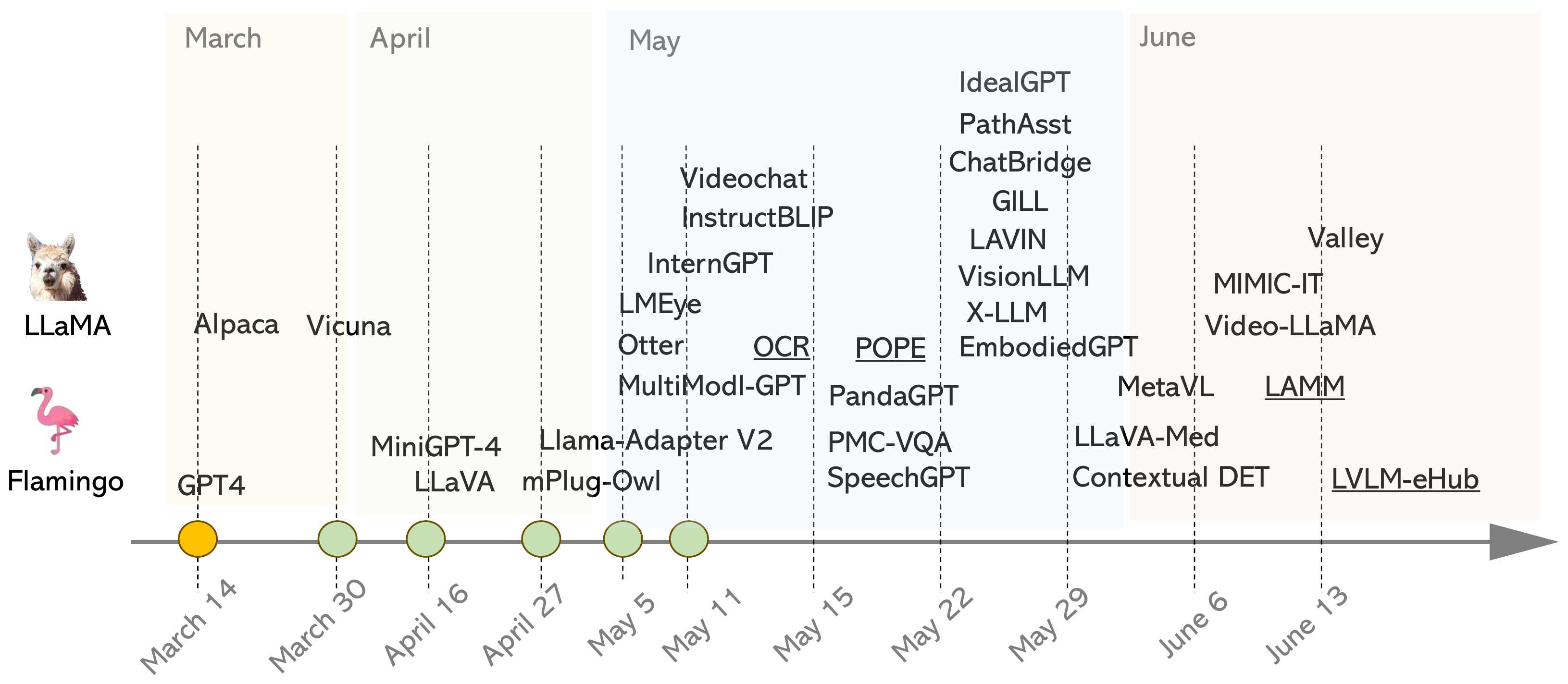} \\
(a) The surge of papers on LMMs in the past three months: March 14 - June 19, 2023. Those with an underline indicate benchmarks, otherwise indicate models. \vspace{2mm}\\
\includegraphics[width=1.0\textwidth]{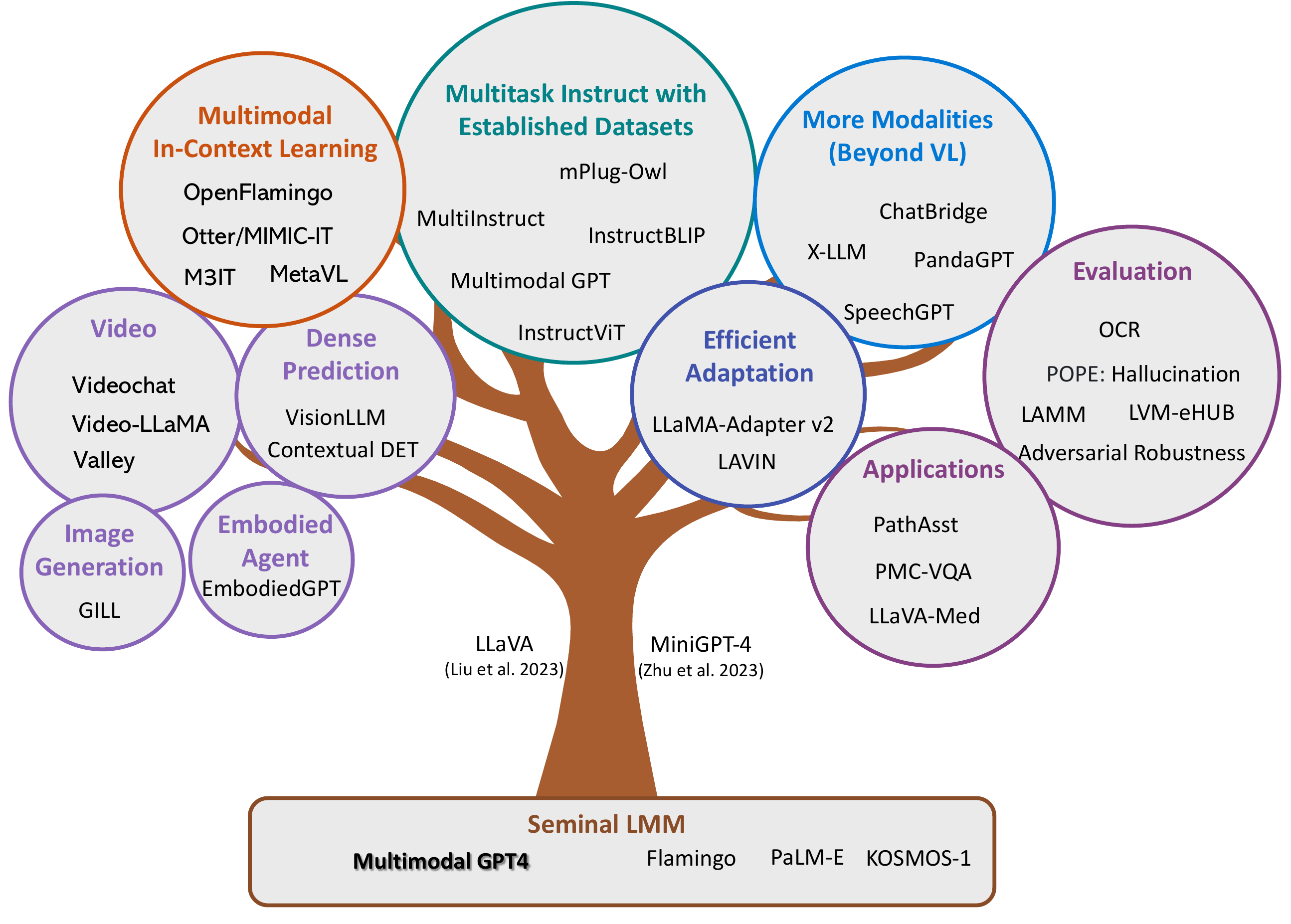} \\
(b) Summary and categorization of papers on LMMs. \\
\end{tabular}
\vspace{-0mm}
\caption{Review and summary for the emerged LMM literature.}
\label{fig:emgerging_topics}  
  \vspace{-1mm}
\end{figure}

The history of recent instructed tuned LMM are illustrated in Figure~\ref{fig:emgerging_topics} (a). 
Due to the popularity of ChatGPT and GPT-4,  instructed tuned LMM appears as an emerging line of research in the past three months after GPT-4 was proposed. Alpaca and Vicuna were proposed to make LLaMA more instruction-following in the language domain in March.
In two weeks, MiniGPT-4 and LLaVA were proposed to make Vicuna to see and chat about the visual world. In ten days, Llama-Adpter v2 and mPlug-OWL started to compare performance with MiniGPT-4/LLaVA, indicating the beginning of model evolution. The data points in April are relatively sparse.
In May, a large number of LMM papers appeared on arXiv, which improve this line of research from many different aspects. The momentum is till going in June.

It is easy to lose track of all the recent papers for the readers, so as well in our literature review. To better organize the literature, we group them based on specific research topics in this tutorial, shown in Figure~\ref{fig:emgerging_topics} (b). The early LMMs with billions of parameters include GPT-4~\citep{gpt4}, Flamingo~\citep{alayrac2022flamingo}, 
PaLM-E~\cite{driess2023palm} and KOSMOS-1~\citep{huang2023language}. In constrast to these proprietary LMMs, LLaVA/MiniGPT-4 open the opportunities to build LMMs with open-source resource.
We will discuss the several topics as below, in addition to dense prediction~\citep{wang2023visionllm,zang2023contextual}, video~\citep{zhang2023video,luo2023valley,li2023videochat}, image generation~\citep{koh2023generating} and embodied agent~\citep{mu2023embodiedgpt}.

\subsubsection{More Modalities (Beyond VL)}
\begin{itemize}[leftmargin=7.5mm]
\setlength{\itemsep}{2pt}
\footnotesize
\item 
{\it ChatBridge: Bridging Modalities with Large Language Model as a Language Catalyst~\citep{zhao2023chatbridge}}

\item
{\it PandaGPT: One Model To Instruction-Follow Them All~\citep{su2023pandagpt}}

\item
{\it SpeechGPT: Empowering large language models with intrinsic cross-modal conversational abilities~\citep{zhang2023speechgpt}}

\item
{\it X-LLM: Bootstrapping Advanced Large Language Models by Treating Multi-Modalities as Foreign Languages~\citep{chen2023x}}
\end{itemize}

\begin{figure}[h!]
\centering  
\vspace{-0mm}
\includegraphics[width=0.99\textwidth]{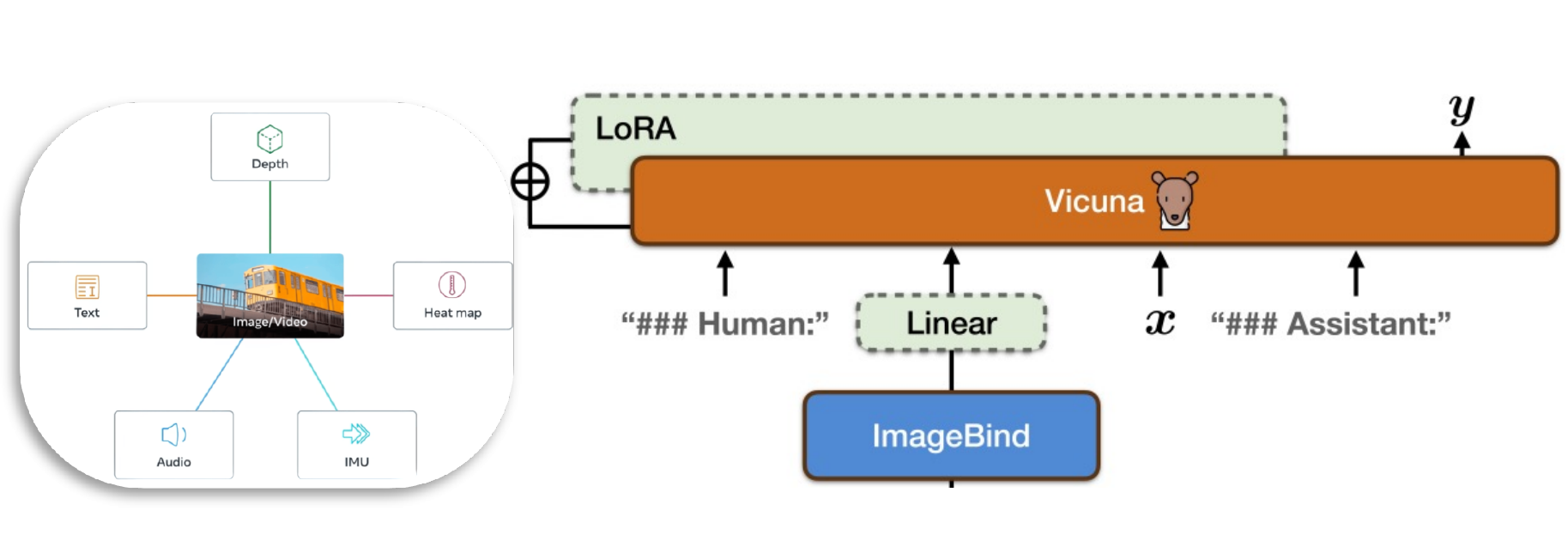} 
\includegraphics[width=0.8\textwidth]{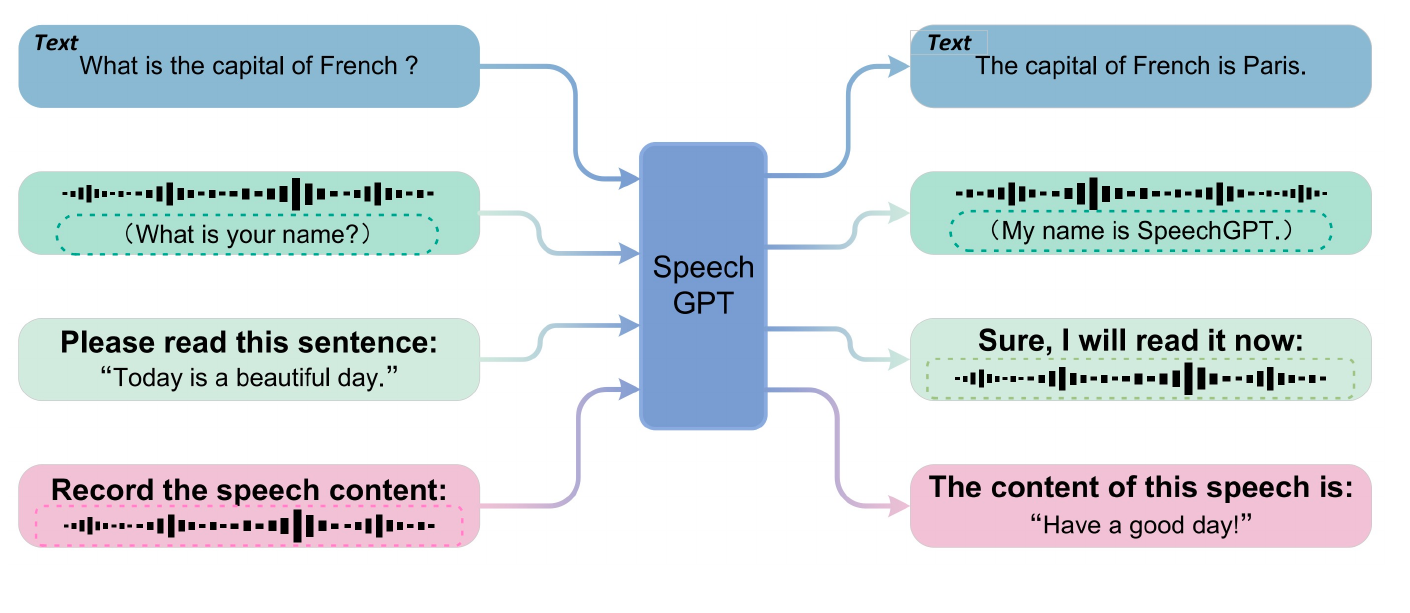} \\
\vspace{-0mm}
\caption{LLMs with more modalities. Top: PandaGPT that enables other modalities via ImageBind; Bottom: SpeechGPT that allows language and speech in both input and output. Image credits from~\citep{su2023pandagpt,zhang2023speechgpt}.}
\label{fig:more_modalities}  
  \vspace{-1mm}
\end{figure}

While LMM extends LLM by adding the vision modality into language, it is natural to further extend the framework to include more modalities beyond vision and language. Following this spirit, several attempts have been made. In Figure~\ref{fig:more_modalities}, PandaGPT leverages ImageBind to add more modalities into LMMs. The ImageBind model~\citep{girdhar2023imagebind} learns a single, shared representation space for text, image/video,  audio, sensors that record depth (3D), thermal (infrared radiation), and inertial measurement units (IMU), which calculate motion and position. ImageBind provides a holistic understanding of the visual world that connects objects in a photo with how they will sound, their 3D shape, how warm or cold they are, and how they move. By training a projection layer for one modality in LMM, the model can zero-shot transfer to infer over other modalities due to the shared multimodal embedding space. Another representative model is SpeechGPT, where language and speech modalities are enabled for both input and output ends. Despite of rich model variations, the idea to connect diverse modalities is similar to LMM that adds images into LLMs.

\subsubsection{Multitask Instruct with Established Academic Datasets/Tasks}
\begin{itemize}[leftmargin=7.5mm]
\setlength{\itemsep}{2pt}
\footnotesize
\item 
{\it MultiInstruct: Improving Multi-Modal Zero-Shot Learning via Instruction Tuning~\citep{xu2022multiinstruct}}

\item
{\it mPlug-OWL: Modularization empowers large language models with multimodality~\citep{ye2023mplug}}

\item
{\it InstructBLIP: Towards general-purpose vision-language models with instruction tuning~\citep{dai2023instructblip}}

\item
{\it Multimodal-GPT: A vision and language model for dialogue with humans~\citep{gong2023multimodal}}

\item
{\it Instruction-ViT: Multi-Modal Prompts for Instruction Learning in ViT~\citep{xiao2023instruction}}
\end{itemize}

As discussed earlier in Section~\ref{sec:instruct_tuning_llm}, instruction tuning in the language domains is implemented in two different ways: fine-tuning the model on a wide range of tasks using human-annotated prompts and feedback~\citep{ouyang2022training}, or supervised fine-tuning using public benchmarks and datasets augmented with manually or automatically generated instructions~\citep{wang2022benchmarking}. The former is good at user-oriented daily life tasks, and the latter is good at achieving good numbers on established benchmarks. LLaVA/MiniGPT-4 can be categorized as the former class. Several other works  either target for the latter class or combine both classes.

\begin{figure}[t!]
\centering  
\vspace{-0mm}
\includegraphics[width=1.00\textwidth]{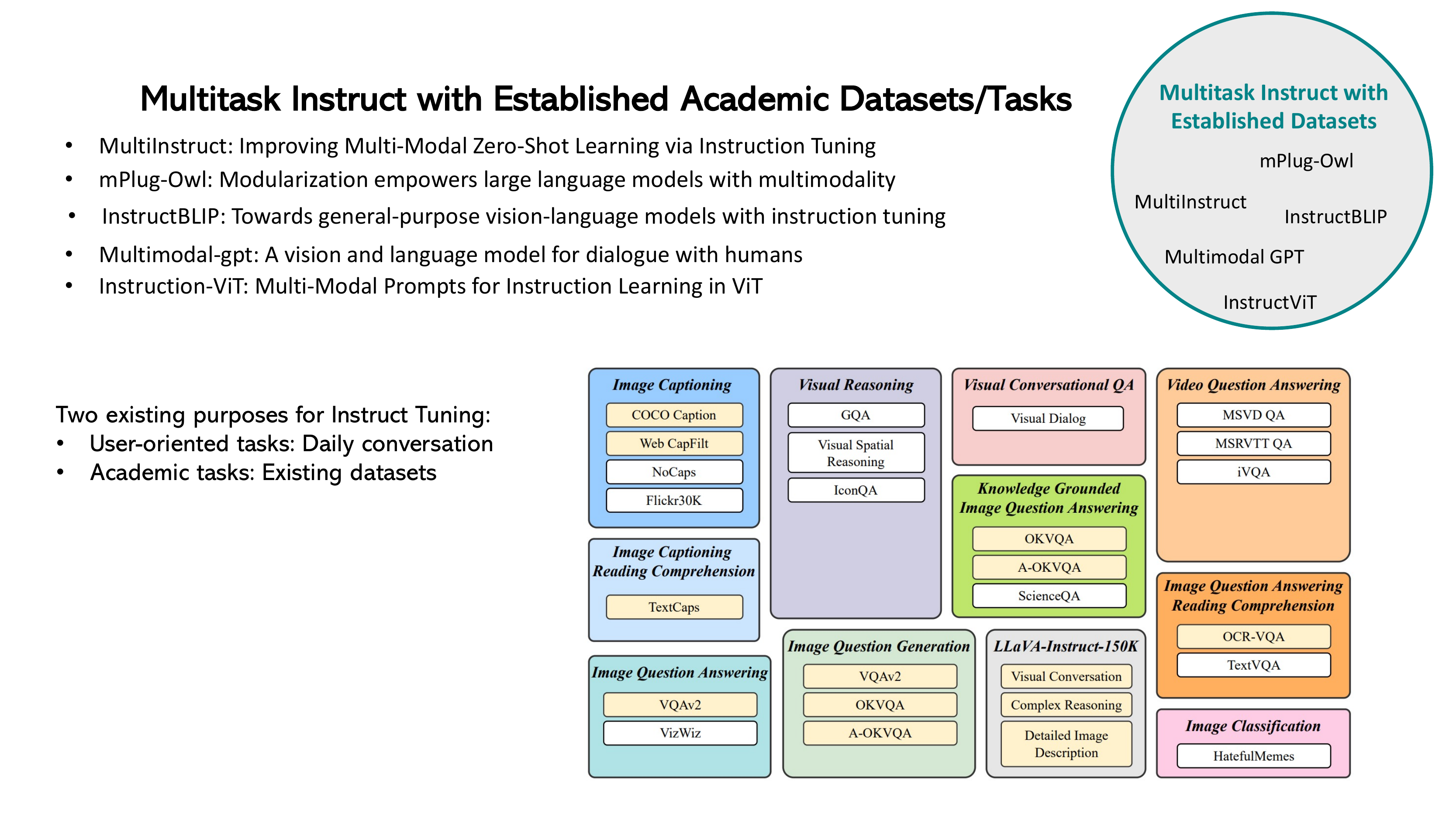} \\
\vspace{-0mm}
\caption{The vision-language tasks covered in InstructBLIP. Image credits from~\citep{dai2023instructblip}.}
\label{fig:instructblip_tasks}  
  \vspace{-1mm}
\end{figure}

For example, MultiInstruct is an early attempt before open-source LLaMA for instruction tuning with multimodal datasets. 
InstructBLIP is recent work that combine chat and benchmark instruction-following data.
As shown in Figure~\ref{fig:instructblip_tasks}, there are 26 publicly available datasets, covering a wide variety of tasks and capabilities, and transform them into instruction tuning format. Trained on 13 held-in datasets, InstructBLIP attains SoTA zero-shot performance across all 13 held-out datasets, substantially outperforming BLIP-2 and larger Flamingo models.

\subsubsection{Multimodal In-Context-Learning}
\begin{itemize}[leftmargin=7.5mm]
\setlength{\itemsep}{2pt}
\footnotesize
\item 
{\it OpenFlamingo~\citep{anas_awadalla_2023_7733589}}

\item 
{\it Otter: A Multi-Modal Model with In-Context Instruction Tuning~\citep{li2023otter}}

\item
{\it M$^3$IT: A Large-Scale Dataset towards Multi-Modal Multilingual Instruction Tuning~\citep{li2023m3it}}

\item
{\it MetaVL: Transferring In-Context Learning Ability From Language Models to Vision-Language Models~\citep{monajatipoor2023metavl}}
\end{itemize}

Similar to the behaviour of LLMs, which can address a language task by processing examples of the task in their text prompt, multimodal in-context-learning refers to an visual and text interface can steer the model towards solving a multimodal task. Given a few example pairs of visual inputs and expected text responses composed in the multimodal prompt, the model can be asked a question with a new image or video, and then generate an answer. 

\begin{figure}[h!]
\centering  
\vspace{-0mm}
\includegraphics[width=1.00\textwidth]{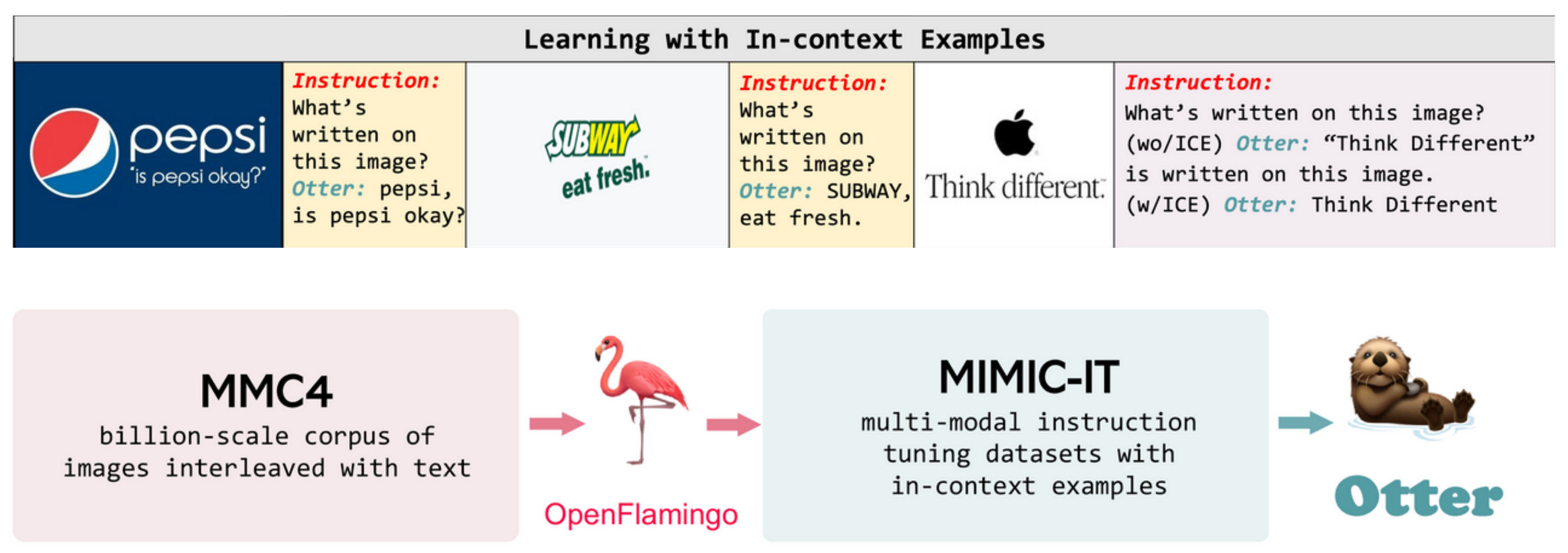} \\
\vspace{-0mm}
\caption{Top: Illustration of multimodal in-context-learning of Otter. Bottom: the training process from OpenFlamingo to Otter. Image credits from~\citep{li2023mimic}.}
\label{fig:multimodal_in_context_learning}  
  \vspace{-1mm}
\end{figure}

OpenFlamingo~\citep{anas_awadalla_2023_7733589} is an open source version of DeepMind's Flamingo model, trained on Multimodal C4 dataset~\citep{zhu2023multimodal}, which is a billions-scale corpus of image interleaved with text. To explicit enhance the multimodal in-context-learning ability of LMMs,  MIMIC-IT~\citep{li2023mimic} dataset is constructed, which is 2.4M multimodal instruction instances with in-context examples. By tuning OpenFlamingo on MIMIC-IT, a new model Otter is obtained with a stronger instruction-following ability. The model life cycle is summarized in Figure~\ref{fig:multimodal_in_context_learning}. Using two image-text pairs as the context, Otter learns the concise answering style demonstrated by the examples, otherwise a tedious response is generated.

\subsubsection{Parameter-Efficient Training}
\begin{itemize}[leftmargin=7.5mm]
\setlength{\itemsep}{2pt}
\footnotesize
\item 
{\it LLaMA-Adapter V2: Parameter-Efficient Visual Instruction Model~\cite{gao2023llama}}

\item
{\it Cheap and Quick: Efficient Vision-Language Instruction Tuning for Large Language Models~\cite{luo2023cheap}}

\item
{\it QLoRA: Efficient Finetuning of Quantized LLMs~\cite{dettmers2023qlora}}
\end{itemize}

While fine-tuning very large models often leads to high performance, it is prohibitively expensive; For example, regular 16-bit fine-tuning of a LLaMA 65B parameter model~\citep{touvron2023llama}  requires more than 780 GB of GPU memory~\cite{dettmers2023qlora}. Therefore, it is critical to reduce the memory footprint of LLMs/LMMs, especially when it comes to improve the accessibility of large models to a wider community. 

\begin{figure}[h!]
\centering  
\vspace{-0mm}
\includegraphics[width=1.00\textwidth]{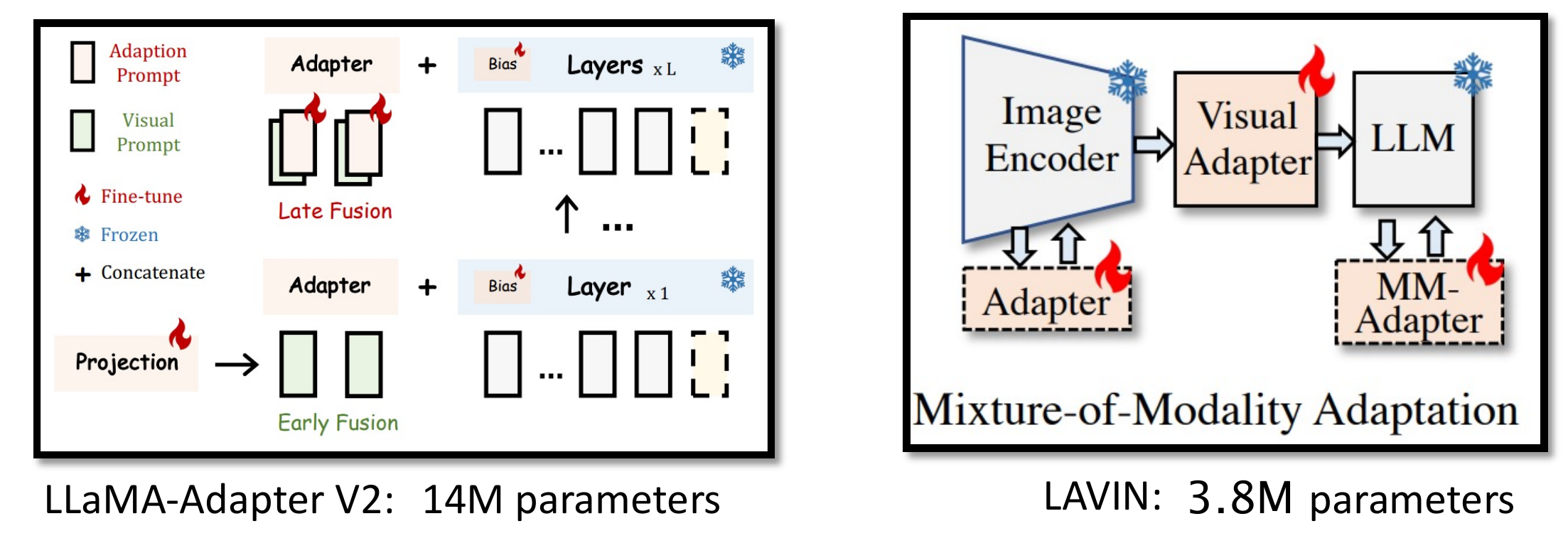} \\
\vspace{-0mm}
\caption{Parameter-efficient methods for LMMs. Image credits from~\citep{gao2023llama,luo2023cheap}.}
\label{fig:parameter-efficient}  
  \vspace{-1mm}
\end{figure}

Parameter-efficient training is an effective approach for LMM adaptation. Two representative methods are illustrated in Figure~\ref{fig:parameter-efficient}. They freeze most of the model parameters, and only allow a small of trainable parameter to update with domain specific data. For example, LLaMA Adapter v2 and LAVIN only has 14M and 3.8M trainable parameters, compared with 7B/13B LLM parameters.
Another efficient training method is quantization. The recent QLoRA finetunes 65B LLaMA for 24 hours on a single GPU, reaching 99.3\% of the performance level of ChatGPT. Since instruction tuning typically involves a small amount of data, it makes parameter-efficient training or model quantization feasible with limited GPU resources.

\subsubsection{Benchmarks}
\begin{itemize}[leftmargin=7.5mm]
\setlength{\itemsep}{2pt}
\footnotesize
\item 
{\it On the Hidden Mystery of OCR in Large Multimodal Models~\cite{liu2023hidden}}

\item 
{\it Evaluating Object Hallucination in Large Vision-Language Models~\cite{li2023evaluating}}

\item
{\it On Evaluating Adversarial Robustness of Large Vision-Language Models~\citep{zhao2023evaluating}}

\item
{\it LAMM: Language-Assisted Multi-Modal Instruction-Tuning Dataset, Framework, and Benchmark~\citep{yin2023lamm}}

\item
{\it LVLM-eHub: A Comprehensive Evaluation Benchmark for Large Vision-Language Models~\cite{xu2023lvlm}}

\end{itemize}

While LMMs have shown excellent visual recognition and reasoning in an open-set manner with free-form text in many scenarios,
the evaluation of LMMs is becoming an urgent and challenging problem. Several related benchmarks have been developed to evaluate various aspects of LMMs, ranging from their specific  abilities including OCR\cite{liu2023hidden}, object hallucination~\cite{li2023evaluating} and adversarial robustness~\citep{zhao2023evaluating}, to comprehensive evaluation~\citep{yin2023lamm,xu2023lvlm}.

\begin{figure}[t!]
\centering  
\vspace{-0mm}
\includegraphics[width=1.00\textwidth]{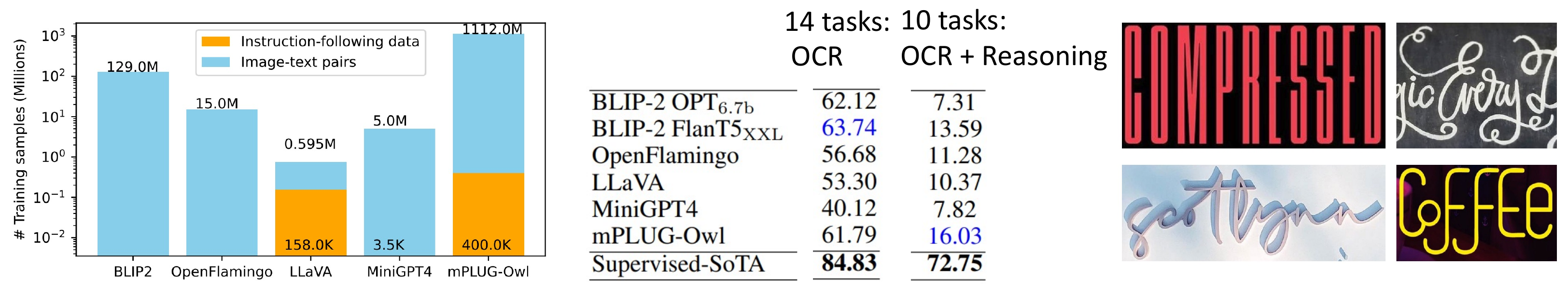} \\
\vspace{-0mm}
\caption{Zero-shot OCR performance of LMMs on 24 datasets. Image credits from~\citep{liu2023hidden}.}
\label{fig:ocr_bench}  
  \vspace{-1mm}
\end{figure}

It is surprising that LMMs shows strong zero-shot OCR performance in the wild, without explicitly training on text recognition data. To shed light on the hidden mystery of OCR in LMMs, a comprehensive empirical study is conducted in~\cite{liu2023hidden} to compare open-source LMMs on 24 academic text recognition datasets, shown in Figure~\ref{fig:ocr_bench}. Three observations are highlighted: (1) LLaVA consistently outperforms miniGPT-4 on 21 out of 24 datasets, despite LLaVA being trained with an order of magnitude smaller training data. (2) Training with significantly larger training data leads to higher OCR performance, as demonstrated by BLIP2~\citep{li2023blip} and mPLUG-Owl. (3) In most cases, supervised SoTA results significantly outperform zero-shot LMM. However, it is worth noting that in the WordArt dataset~\citep{xie2022understanding}, which primarily features challenging artistic text, BLIP2 surpasses supervised SoTA. This reveals the potential of LMM in recognizing more complex text types.

\begin{figure}[h!]
\centering  
\vspace{-0mm}
\includegraphics[width=1.00\textwidth]{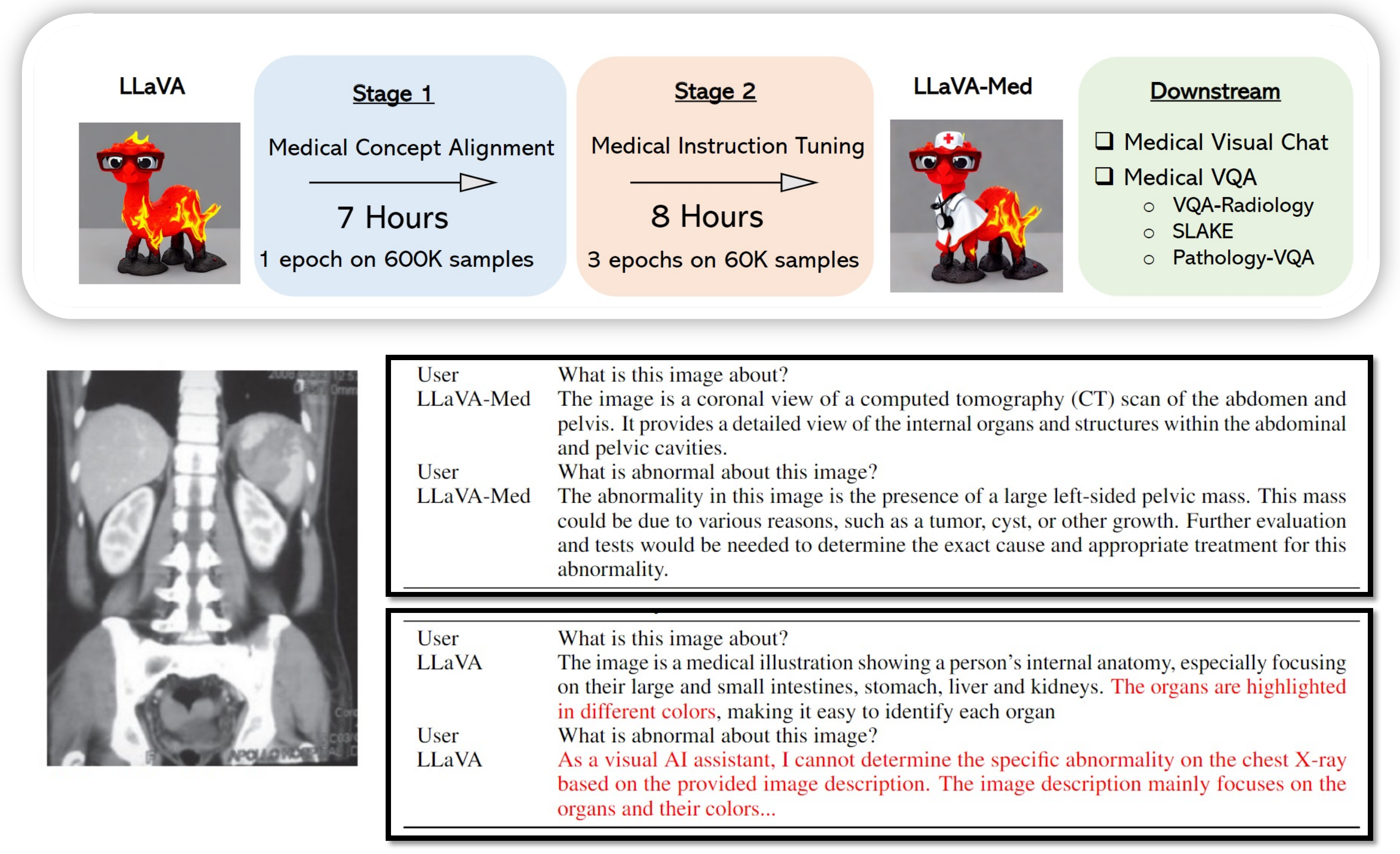} \\
\vspace{-0mm}
\caption{Application of LMMs to bio-medicine. Top: The domain adaptation from LLaVA to LLaVA-Med. Bottom: The chat behaviors of two chatbots. Image credits from~\citep{li2023llava}.}
\label{fig:llava_med}  
  \vspace{-1mm}
\end{figure}

\subsubsection{Applications}

\begin{itemize}[leftmargin=7.5mm]
\setlength{\itemsep}{2pt}
\footnotesize
\item 
{\it PathAsst: Redefining Pathology through Generative Foundation AI Assistant for Pathology~\citep{sun2023pathasst}}

\item
{\it PMC-VQA: Visual Instruction Tuning for Medical Visual Question Answering~\citep{zhang2023pmc}}

\item
{\it LLaVA-Med: Training a Large Language-and-Vision Assistant for Biomedicine in One Day~\citep{li2023llava}}
\end{itemize}

The success of ChatGPT/GPT-4 in the general domain has inspired the interests in building assistants in the vertical domains such as medicine, gaming and education. Such domain-specific assistants can have the several advantages over the general domain counterpart: (1) training high-quality domain knowledge makes the assistants more helpful, (2) the model size can be smaller, and thus severing cost is low, (3) the sensitive user prompt data can be maintained internally by serving the model at local, and the privacy issue can be avoided.

LMMs have been recently explored in the biomedical domain~\citep{sun2023pathasst,zhang2023pmc,li2023llava}, where conversational generative AI has demonstrated remarkable promise for empowering biomedical practitioners.
LLaVA-Med is a cost-efficient approach for training a vision-language conversational assistant that can answer open-ended research questions of biomedical images. The key idea is to leverage a large-scale, broad-coverage biomedical figure-caption dataset extracted from PubMed Central, use GPT-4 to self-instruct open-ended instruction-following data from the captions, and then fine-tune a large general-domain vision-language model LLaVA using a novel curriculum learning method. Specifically, the model first learns to align biomedical vocabulary using the figure-caption pairs as is, then learns to master open-ended conversational semantics using GPT-4 generated instruction-following data, broadly mimicking how a layperson gradually acquires biomedical knowledge. In Figure~\ref{fig:llava_med}, we provide examples on the biomed visual conversations of different chatbots. LLaVA-Med precisely answers the questions with biomedical knowledge, while LLaVA behaves like a layperson, who hallucinate based on commonsense.

\pagebreak
\newpage
\newpage
\section{How Close We Are with OpenAI Multimodal GPT-4?}
\label{conclusions}

With all these new works, are we close or even surpassing OpenAI Multimodal GPT-4? It is encouraging to see that the open-source community has quickly developed a variety of models and prototypes for various new capabilities. For example, LLaVA/Mini-GPT4 paves the way towards building multimodal chatbots, with some examples that reproduce the results in OpenAI GPT-4 technique report; GILL~\citep{koh2023generating} extends LMMs for end-to-end image generation, to our best knowledge, this is a capability that the current GPT-4 does not exhibit. From the perspective of enabling new multimodal capabilities with the minimum prototypes, the open-source community seems close to OpenAI Multimodal GPT-4, by exploring the baby steps towards building the general-purpose multimodal assistant.

\begin{figure}[h!]
\centering  
\vspace{-0mm}
\includegraphics[width=1.00\textwidth]{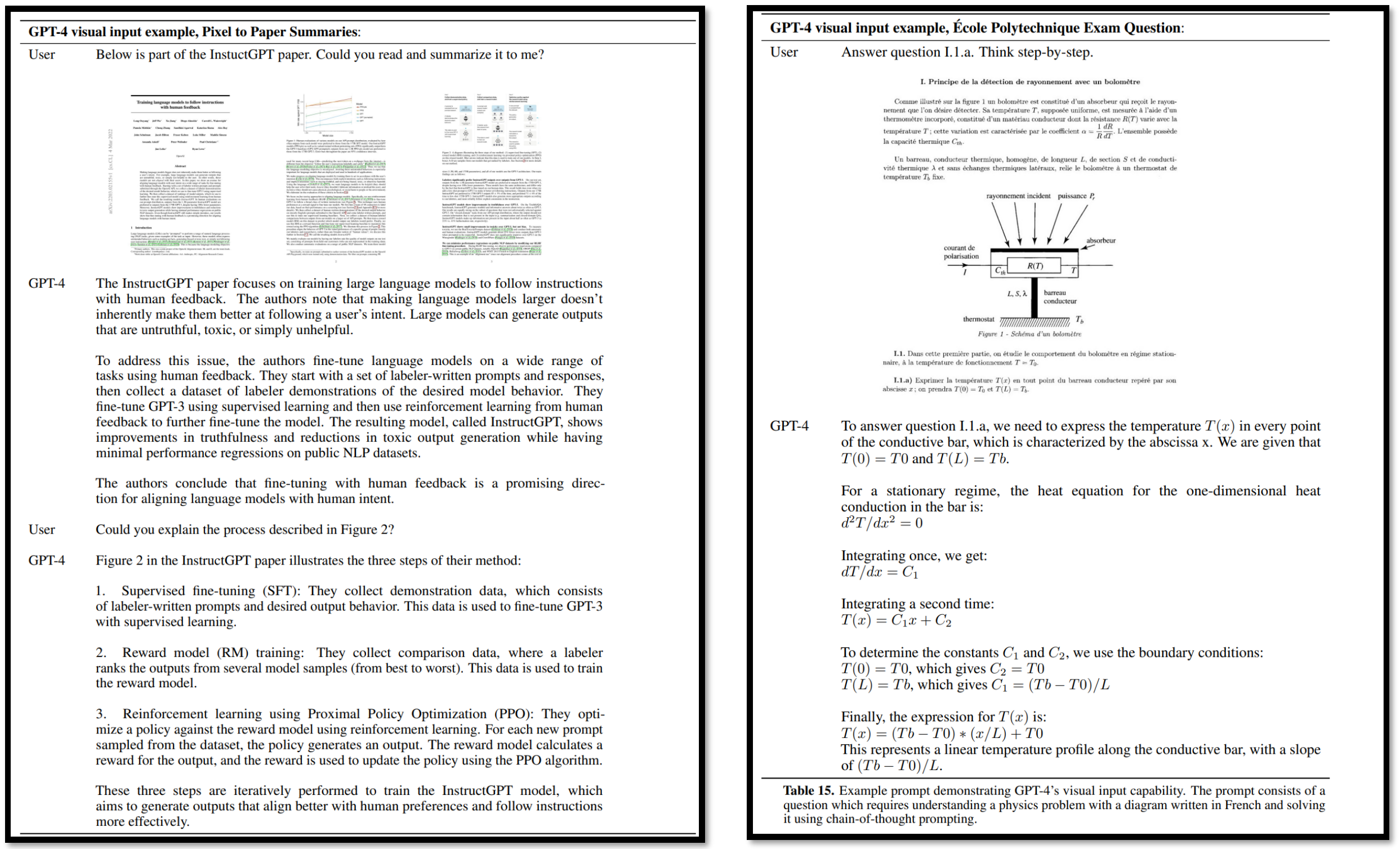} \\
\vspace{-0mm}
\caption{Challenging visual examples from OpenAI technique report~\cite{gpt4} that requires scaling the visual understanding and reasoning capability.}
\label{fig:gpt4_examples_gap}  
  \vspace{-1mm}
\end{figure}

However, there is a large gap in terms of scaling a given capability, for example, even the for visual reasoning capability that we have observed in LLaVA. Figure~\ref{fig:gpt4_examples_gap} shows two more visual examples from OpenAI technique report. To correctly answer the questions, it requires models to understand multiple high-resolution images and long sequence, as well we responding with  domain knowledge. It requires much larger compute and more powerful language models, which are not available for most people.

In summary,  we have presented the background and strong capabilities of large multimodal models, reviewed instruction tuning in LLMs, and showed how we can build a prototype such as LLaVA and minigpt4 using open-sourced resources. We also summarize and cateorized the most recent papers merged on this line of research to help thoese who are interested to gain the momentum to start the journey of LMM research.

To discuss the next steps to work on as a community, one sustainable suggestion can be that those with resource can continue focusing on the scaling success and study new emerging properties, while others focus on prototypes for new functionalities and evaluation, as well as developing techniques to reduce the compute barriers and thus allow more accessibility for larger model compute.

\newpage
\begin{ack}
We thank all authors who have contributed to the related papers in LLM/LMM, which makes the tutorial possible. We have tried to track related papers for the CVPR tutorial before June 19, 2023, but may not cover all the papers on the topic, due to the fast research pace in LMMs. Apologies in advance.

\end{ack}

\bibliography{egbib}
\bibliographystyle{plain}

%%%%%%%%%%%%%%%%%%%%%%%%%%%%%%%%%%%%%%%%%%%%%%%%%%%%%%%%%%%%

\end{document}